\pdfoutput=1
\UseRawInputEncoding
\documentclass{article}
\usepackage{float}
\usepackage[margin=1.0in]{geometry}
\usepackage{abstract}
\usepackage{afterpage}
\usepackage[backend=bibtex,style=nature,natbib=true,maxbibnames=99,minalphanames=3]{biblatex}
\usepackage{hyperref}

\usepackage{caption}
\usepackage{lmodern}
\usepackage{threeparttable}
\usepackage[dvipsnames,svgnames, table,xcdraw]{xcolor}
\usepackage{relsize}
\usepackage{tcolorbox}
\usepackage{listings}
\usepackage{subcaption}
\usepackage{algorithm}
\usepackage{algorithmic}
\usepackage{comment}
\usepackage{graphicx}
\usepackage{booktabs}
\usepackage{multirow}
\usepackage{todonotes}
\usepackage{enumitem}
\usepackage{url}
\usepackage{mathpazo}
\usepackage{amsmath,amssymb,amsthm,amsfonts,bm}
\usepackage[toc,page,header]{appendix}
\usepackage{fancyhdr}
\usepackage[T1]{fontenc}

\usepackage{auth_detailed}

\counterwithout{figure}{section}
\counterwithout{table}{section}

\definecolor{darkgoldenrod}{rgb}{0.72, 0.53, 0.04}
\definecolor{backgroundcolor}{RGB}{250, 250, 252}
\definecolor{keywordcolor}{RGB}{30, 0, 178}
\definecolor{stringcolor}{RGB}{204, 0, 102}
\definecolor{numbercolor}{RGB}{0, 128, 128}
\definecolor{emphcolor}{RGB}{30, 0, 178}
\definecolor{commentcolor}{RGB}{0, 128, 0}
\definecolor{basiccodecolor}{RGB}{61, 61, 61}

\lstdefinestyle{customstyle}{
    backgroundcolor=\color{backgroundcolor},
    commentstyle=\color{commentcolor},
    keywordstyle=\color{keywordcolor},
    numberstyle=\color{numbercolor},
    stringstyle=\color{stringcolor},
    basicstyle=\color{basiccodecolor}\ttfamily\footnotesize,
    breakatwhitespace=false,
    breaklines=true,
    captionpos=b,
    keepspaces=true,
    numbers=left,
    numbersep=5pt,
    xleftmargin=2em,
    xrightmargin=2em,
    showspaces=false,
    showstringspaces=false,
    showtabs=false,
    tabsize=1,
    frame=single,
    framesep=5pt,
    framexleftmargin=1.5em,
    framexrightmargin=1.5em,
    framextopmargin=1pt,
    framexbottommargin=1pt,
    aboveskip=10pt,
    belowskip=10pt,
    breakautoindent=true,
    emphstyle={\color{emphcolor}},
    extendedchars=true,
}

\usepackage[normalem]{ulem}
\usepackage{xcolor}

\newcommand{\scrcmd}{\textsuperscript}
\newcommand{\scr}[1]{\scrcmd{#1}}

\definecolor{logocolor}{RGB}{30, 0, 178}
\definecolor{darkerlogocolor}{RGB}{20, 0, 145}
\newtcolorbox{ttcolorbox}[1][]{colframe=darkerlogocolor, colback=darkerlogocolor!4!white, title=#1}
\newtcolorbox{apxtcolorbox}[1][]{colframe=black, colback=black!3!white, title=#1}

\usepackage{fontawesome5}
\usepackage{helvet}
\usepackage{xspace}
\DeclareRobustCommand{\corrAuthor}{\textsuperscript{\faEnvelope[regular]}}

\hypersetup{
  colorlinks=true,
  urlcolor=blue,
  citecolor=ForestGreen,
  linkcolor=blue
}
\definecolor{echokeplerblue}{HTML}{6352a3}

\newcommand{\arialtitle}[1]{{\fontfamily{phv}\selectfont #1}}
\def\ourmethod{EchoKepler\xspace}

\title{{\fontsize{15pt}{18pt}\selectfont \textbf{\arialtitle{An agentic system for automated evidence-grounded echocardiography interpretation}}}}

\author{\name Hanwen Xu$^{1,*,}$\corrAuthor,
\name Patricia Xiao$^{1,*}$,
\name Xinyue Liang$^{1,*}$,
\name David Zhang$^{2}$,
\name Shin Lin$^{4,}$\corrAuthor,
\name Jay Pal$^{3,}$\corrAuthor,
\name Sheng Wang$^{1,}$\corrAuthor\\\newline
$^{1}$ School of Computer Science and Engineering, University of Washington, Seattle, WA\\
$^{2}$ Virginia Mason Franciscan Health, Tacoma, WA\\
$^{3}$ Department of Surgery, School of Medicine, University of Washington, Seattle, WA\\
$^{4}$ Emory University, Division of Cardiology, Atlanta, GA\\
\newline
*~~~Equal contribution\\
}

\begin{document}

\fancyhead[L]{\textsf{\textcolor{echokeplerblue}{EchoKepler}}}
\fancyhead[R]{An agentic system for echocardiography interpretation}
\setlength{\headheight}{13pt}
\pagestyle{fancy}

\newcounter{suppfigure}
\newcounter{supptable}
\makeatletter
\newcommand\suppfigurename{Supplementary Figure}
\newcommand\supptablename{Supplementary Table}
\newcommand\suppfigureautorefname{\suppfigurename}
\newcommand\supptableautorefname{\supptablename}
\let\oldappendix\appendix
\renewcommand\appendix{%
    \oldappendix
    \setcounter{figure}{0}%
    \setcounter{table}{0}%
    \renewcommand\figurename{\suppfigurename}%
    \renewcommand\tablename{\supptablename}%
}
\makeatother

\maketitle

\renewcommand{\abstractnamefont}{\normalfont\normalsize\bfseries}
\renewcommand{\abstracttextfont}{\normalfont\normalsize}

\renewcommand{\abstractname}{Abstract}
\begin{abstract}
\abstracttextfont
Echocardiography is performed millions of times each year to monitor valvular disease, heart failure and ventricular function. Its interpretation is a demanding multi-step process requiring years of expertise and practice. The analysis is fundamentally challenging because each study comprises over a hundred videos acquired from multiple standardized views. Clinicians must gather decisive evidence across acquisition views, assess structural and functional abnormalities, and decide the next examination before making a diagnosis. Yet existing foundation models rely on single-step predictions, falling short of this iterative, multi-step reasoning.

Here, we introduce \ourmethod, an agentic system for automated, evidence-grounded echocardiography interpretation. Inspired by how cardiologists read echos, \ourmethod orchestrates four core steps: acquisition view prediction, multi-turn tool invocation, self-reflection, and report generation. \ourmethod selectively invokes a suite of 24 diagnostic tools, customizing tool choice and execution order on a per-patient basis. We evaluate \ourmethod extensively on held-out cohorts from UW Medicine. Its diagnostic tools outperform the strongest foundation model, EchoCLIP, by 9.6\% in average AUROC across 24 detection tasks. For clinical decision-making, \ourmethod predicts surgical indications from echocardiograms alone with an AUROC of 0.958. As an agentic reporting system, it substantially outperforms strong general-purpose and medical-specific models including GPT-5.4 and MedGemma v1.5. Patient-specific selective tool invocation also decisively surpasses executing all available tools. Finally, enabled by its self-reflective design, \ourmethod examines interval changes across serial studies, enabling longitudinal analysis of disease progression. \ourmethod thus delivers accurate, evidence-grounded, and traceable interpretations of echocardiograms.

\end{abstract}

\section{Introduction}
In cardiovascular medicine, echocardiography is widely used and accessible~\cite{papolos2016us}, with millions of studies undertaken each year~\cite{virnig2014trends, colebourn2024state} to diagnose and monitor valvular diseases, heart failure, etc~\cite{bonow1998guidelines, heidenreich20222022}. Yet this accessibility comes at the cost of complexity~\cite{mitchell2019guidelines, zhang2018fully}. Each echocardiographic study comprises dozens of videos acquired from multiple standardized views, and substantial effort is needed to interpret them by reviewing findings of structure and function. Because clinically decisive evidence typically appears in only a small subset of over dozens to a hundred videos in a single study, clinicians must navigate a large and heterogeneous set of inputs~\cite{huang2023detecting, vrudhula2024high}. This process is very time consuming and requires years of specialized training and practice.

In real-world practice, it is often impossible for cardiologists to reach a complete diagnosis in a single glance. Evidence is gathered over multiple steps~\cite{nagueh2016recommendations, zoghbi2017recommendations}, interpreting findings in each view when assessable, and deciding which aspect to examine next before arriving at a conclusion. If longitudinal examinations exist, serial studies are then compared to judge how a patient's condition has evolved and when clinical interventions are appropriate. Evidence-driven, stepwise and selective reasoning are central to how experts read echocardiograms, and a clear chain of evidence is essential for an accurate interpretation.

While biomedical AI progresses rapidly, it still falls short of this standard. Current biomedical vision-language models generate entire clinical reports in a single pass~\cite{zambrano2025clinically, li2023llava, zhang2024generalist, hamamci2026generalist, chao2025evaluating, tu2024towards}.  This strategy is prone to hallucinating findings without clear evidence in the images. In echocardiography, foundation models have been proposed with the idea of providing generalizable visual features to support various downstream tasks. The scaling of data and model size have yielded promising results to date. From EchoCLIP~\cite{christensen2024vision} to EchoPrime~\cite{vukadinovic2026comprehensive}, the pretraining data size is ten times larger, and the analysis has been extended from single view to multi-view examinations. Despite the exciting progress, their capabilities in performing end-to-end interpretation, such as echo report generation, still remain largely unexplored. With the rise of agentic models, AI systems are engineered to accomplish complex tasks, such as coding~\cite{jiang2025adaptation, yang2024swe, wang2024survey}. At their core, these systems operate through executing actions via tools, observing environmental feedback, and iteratively deciding the next step until task completion~\cite{sun2025scaling}. Instead of relying on a single forward pass, agentic systems decompose tasks into verifiable steps and condition subsequent actions on real-time evidence.

Here, we introduce \ourmethod, an agentic system for automated, evidence-grounded
echocardiogram interpretation.  We equip \ourmethod with a suite of 24 diagnostic
tools, each predicting whether a specific abnormality is observed from the relevant echo views. \ourmethod orchestrates four core steps: a view prediction step that determines the video acquisition view, a multi-turn tool-calling step that gathers diagnostic evidence, a self-reflection step that verifies tool predictions, and a report generation step that articulates the verified findings and conclusions. Crucially, instead of following a fixed tool-use pipeline, \ourmethod customizes the choice and execution schedule of diagnostic tools on a per-patient basis. Both the agent and tools are supported by an echocardiography vision foundation model pretrained on 30 million frames, thereby grounding all decisions and tool invocations in rich visual representations. We optimize the entire architecture end-to-end by unifying supervised fine-tuning and reinforcement learning to induce optimal evidence-gathering trajectories.

We validate \ourmethod on held-out cohorts from UW Medicine across a broad range of tasks. First, its diagnostic tools outperform the strongest foundation model, EchoCLIP, by 9.6\% in average AUROC across 24 detection tasks. For clinical decision-making, \ourmethod predicts surgical indications from echocardiograms alone with an AUROC of 0.958, surpassing EchoPrime by a large margin.
As an agentic reporting system, \ourmethod substantially outperforms frontier baselines including GPT-5.4~\cite{openai2026gpt54} and MedGemma v1.5~\cite{sellergren2026medgemma}. Furthermore, ablation experiments confirm that selectively invoking tools on a per-patient basis consistently surpasses naively executing the entire tool suite.
Finally, we explore the clinically critical yet challenging task of longitudinal reporting, interpreting interval changes across serial echo studies. Enabled by its self-reflection step, \ourmethod can deliver verifiable interpretations that faithfully capture temporal disease progression.

\section{Results}
\label{sec:benchmarks}
\begin{figure}[htbp]
  \centering
  \includegraphics[width=\linewidth]{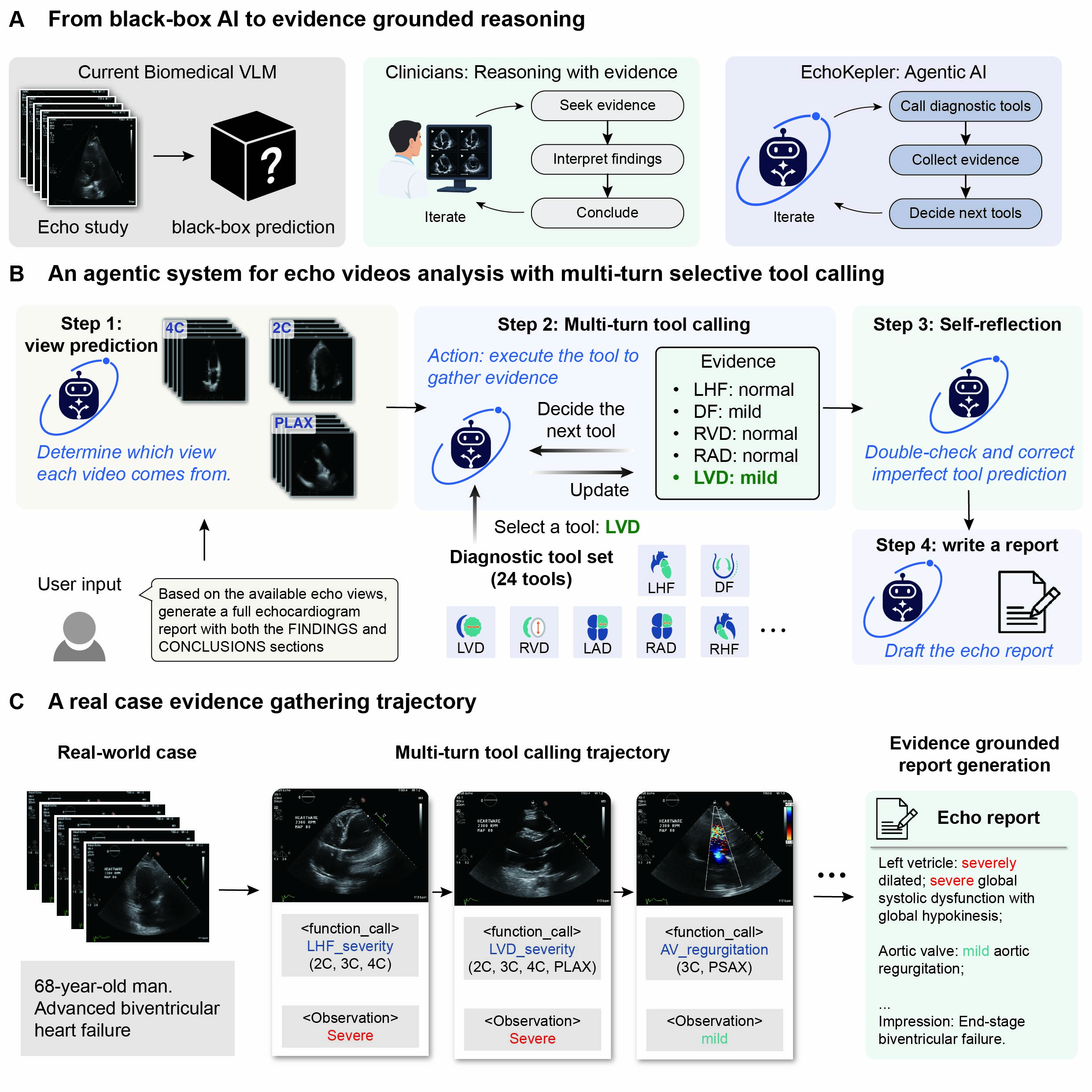}
  \caption{\textbf{Overview of the agentic model \ourmethod for evidence-grounded echocardiographic interpretation. } \textbf{A, }Illustration of the transition from the blackbox AI to evidence-grounded reasoning. \textbf{B, }An overview of the agentic system for echo video analysis. Given a user request to generate a full echocardiographic report, the view prediction step first assigns each input video to its acquisition view (e.g., 4C, 2C, PLAX). The multi-turn tool calling step then operates in a closed loop: at each turn it selects a tool from a set of 24 diagnostic tools, executes it on the relevant views to gather evidence, updates the accumulated evidence pool, and decides which tool to call next. Once evidence gathering is complete, a self-reflection step double-checks the accumulated predictions and corrects imperfect tool outputs, and a report generation step drafts the final echocardiography report from the verified evidence. \textbf{C, }Example trajectory for a 68-year-old man with advanced biventricular heart failure. }
  \label{fig:1}
\end{figure}

\subsection{\ourmethod generates a trace of evidence readout for echocardiography interpretation}

Existing biomedical vision-language models (VLMs) map an entire echo study to a report in a single black-box forward pass. However, in clinical practice, cardiologists actively gather evidence over multiple steps, interpret each finding, and decide which aspect of the study to examine next before reaching a conclusion. \ourmethod emulates this process through a loop of multi-turn diagnostic tool invocation. At each turn, it invokes a tool to evaluate a suspected abnormality, dynamically selecting subsequent tools to accumulate evidence until reaching a confident conclusion. Instead of naively invoking the entire tool set, EchoKepler customizes the choice and execution order of diagnostic tools on a per-patient basis.

The agentic system starts with a series of videos from an echo study and the user's request to generate a full report. The view prediction step uses a classifier trained on view annotations from experienced cardiologists, which determines the acquisition view of each video. 
This step is essential because cardiac abnormalities are assessable only in specific acquisition views, providing the necessary context to guide downstream tool execution.
Next, the multi-turn tool-calling step leverages a specialized suite of 24 diagnostic tools to assess abnormalities spanning valvular heart disease, chamber enlargement, ventricular dysfunction, and pericardial effusion.
At each turn, the agentic system invokes a single diagnostic tool, restricting its input strictly to videos with predicted views relevant to the suspected abnormality.
Upon execution, the returned severity score is integrated into the accumulated evidence as a novel observation, updating the agent's state to guide subsequent tool selection.
Once the accumulated evidence provides sufficient diagnostic certainty, the agent terminates the tool-calling loop and proceeds to report generation. 
To safeguard against potential predictive noise or tool-level inaccuracies, the framework incorporates a self-reflection step that systematically verifies the compiled evidence and rectifies initial interpretations. Finally, the agent synthesizes these validated findings into a coherent clinical report.
In \textbf{Figure~\ref{fig:1}C}, we present a real case showing the trajectory of gathering evidence from \ourmethod and the generation of reports.

To ground decision-making and tool invocations in faithful visual representations, we pretrain a domain-specific foundation model as the visual backbone for the multimodal LLM.
We pretrained a large Vision Transformer (ViT-L) on 30,080,731 echo frames from UW Medicine using DINOv3~\cite{simeoni2025dinov3}, a self-supervised framework that learns high-quality dense visual features. To align the encoder's embedding space with that of the LLM, we fine-tune it with a contrastive objective over 476,974 paired echocardiogram videos and their corresponding clinical reports. Because this foundation model yields rich and expressive representations, its feature space serves as a strong embedding layer for downstream analysis, allowing us to instantiate the entire diagnostic tool suite by simply attaching lightweight classification heads trained on expert clinical annotations.

\begin{figure}[htbp]
  \centering
  \vspace{-3mm}
  \includegraphics[width=\linewidth]{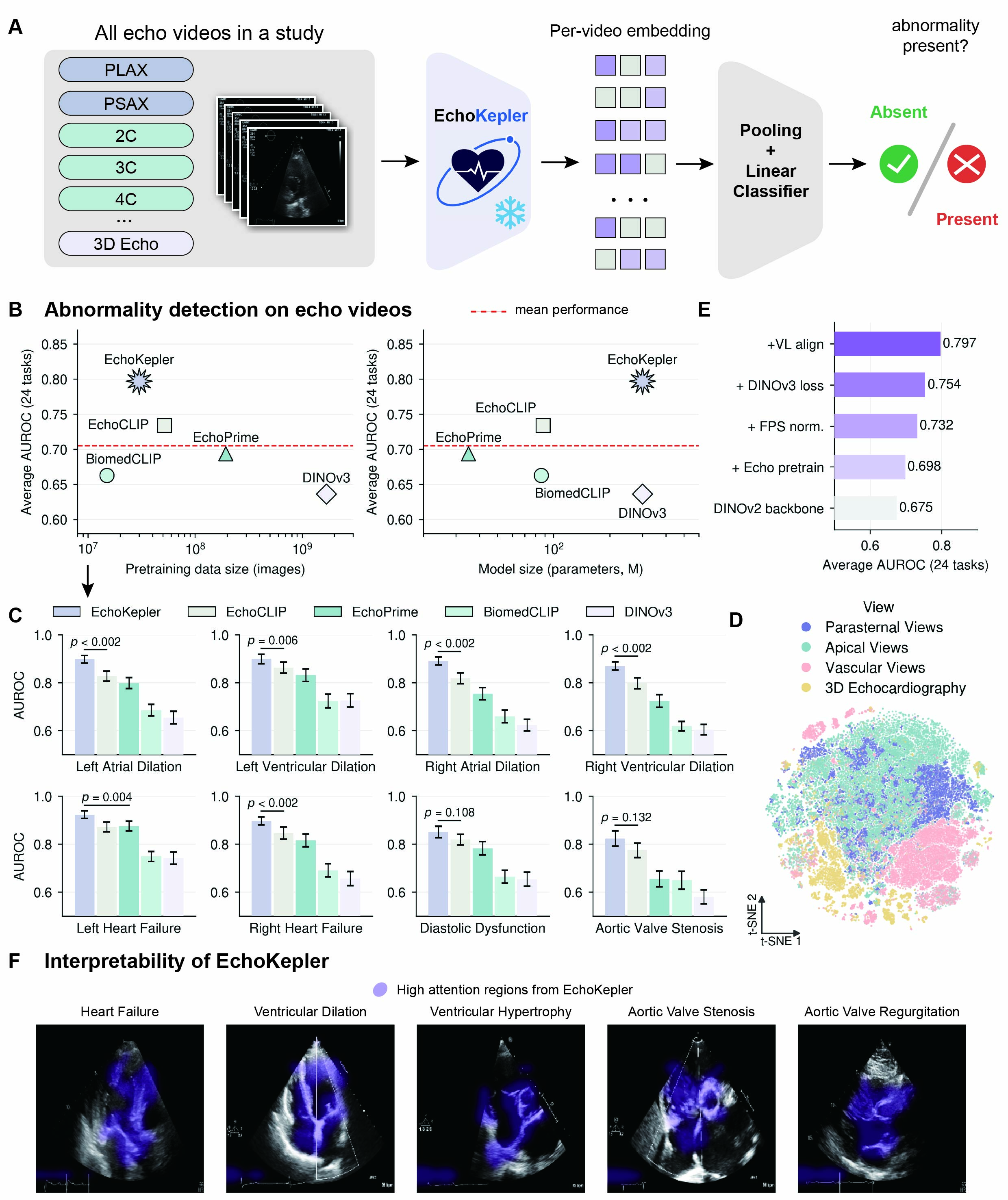}
  \caption{\textbf{\ourmethod gathers evidence via a suite of diagnostic tools.} \textbf{A, } each diagnostic tool uses the \ourmethod encoder to produce per-video embeddings. A classification head was trained to output whether an abnormality was detected. \textbf{B, } Scatter plots comparing the average performance across 24 tasks with the pretraining dataset size and model size of each method. \textbf{C, } Bar plots comparing \ourmethod and competing methods in detecting each abnormality. \textbf{D, } t-SNE plots visualizing the study-level embeddings, color-coded by the type of view. \textbf{E, } Ablation for pretraining on echo videos, FPS normalization across videos, DINOv3 norms, and visual-language alignment. The performances are average AUROC values across 24 tasks. \textbf{F, } Attention maps from EchoKepler for studies across multiple abnormalities. For each case, attention concentrates on the anatomical structure clinically relevant to the finding.}
  \label{fig:2}
\end{figure}

\subsection{A validated suite of tools for gathering echocardiographic evidence}

\begin{figure}[htbp]
  \centering
  \includegraphics[width=0.99\linewidth]{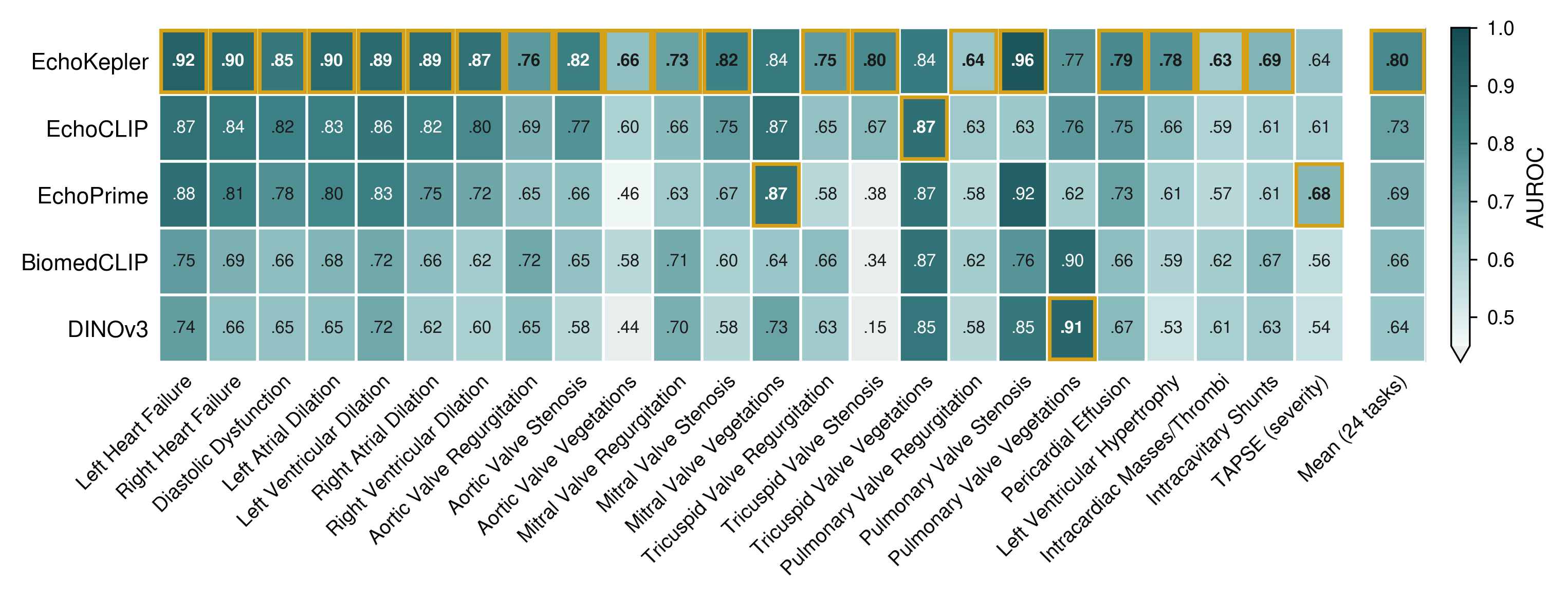}
  \caption{\textbf{Heatmap comparing the AUROC of \ourmethod and baselines for detecting 24 abnormalities.} Each row represents a model and each column represents an individual task. The gold box indicates the best model for each task.}
  \label{sfig:abnormality_auroc}
\end{figure}

\begin{figure}[htbp]
  \centering
  \includegraphics[width=0.99\linewidth]{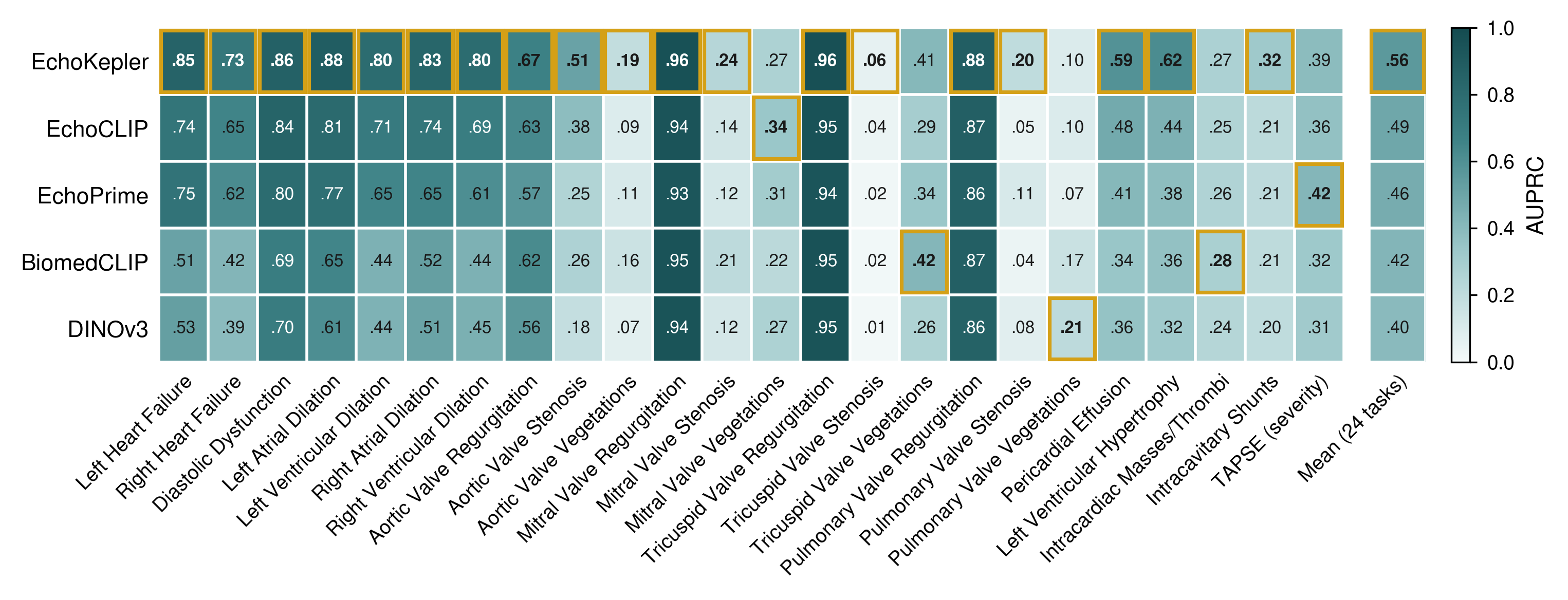}
  \caption{\textbf{Heatmap comparing the AUPRC of \ourmethod and baselines for detecting 24 abnormalities.} Each row represents a model and each column represents an individual task. The gold box indicates the best model for each task.}
  \label{sfig:abnormality_auprc}
\end{figure}

We instantiate a suite of 24 specialized diagnostic tools covering the principal abnormalities routinely assessed in standard echocardiographic evaluations.
Each tool is implemented as a lightweight classification head on top of the shared \ourmethod vision encoder (\textbf{Figure~\ref{fig:2}A}), ensuring low-latency, computationally efficient inference throughout the agentic loop.
To assess their accuracy, we evaluate the tools on a held-out UW Medicine cohort (n=395) across 24 tasks encompassing both structural and functional abnormalities (e.g., chamber dilation, hypertrophy, reduced LV systolic function).
We compare against recent publicly available foundation models, including EchoPrime~\cite{vukadinovic2026comprehensive}, EchoCLIP~\cite{christensen2024vision}, BiomedCLIP~\cite{zhang2025multimodal}, and the general-purpose DINOv3~\cite{simeoni2025dinov3}. \ourmethod substantially outperforms the next best model EchoCLIP by 9.6\% in average AUROC (\textbf{Figure~\ref{fig:2}B}). Remarkably, these gains are achieved without relying on brute-force scaling of model parameters or training data, demonstrating the efficacy of our targeted representation learning framework. When stratifying performance across individual abnormalities, \ourmethod demonstrates consistent improvements, outperforming EchoCLIP on 20 of 24 tasks (\textbf{Figure~\ref{fig:2}C}, \textbf{Supplementary Figures~\ref{sfig:abnormality_auroc},~\ref{sfig:abnormality_auprc}}).

To evaluate the underlying geometry of the \ourmethod embedding space, we visualize two-dimensional t-SNE projections of the embeddings colored by acquisition views (parasternal, apical, vascular, and 3D echocardiography; \textbf{Figure~\ref{fig:2}D}). Despite the encoder never being exposed to view labels, the embedding space naturally organizes by acquisition view, with each category forming a distinct, cohesive cluster.
After confirming the quality of learned representations, we conduct a cumulative ablation study to quantify the contribution of each pretraining component. Starting from an off-the-shelf DINOv2 backbone~\cite{oquab2024dinov2}, we sequentially introduce in-domain echo pretraining, frame-rate normalization, DINOv3 objective formulations, and vision-language alignment. Each design component provides consistent gains, progressively increasing the average AUROC from 0.675 to 0.797 (\textbf{Figure~\ref{fig:2}E}).
Finally, to evaluate how \ourmethod grounds its diagnostic evidence, we examine the interpretability of tool predictions by visualizing attention maps from the vision encoder. Across representative cases, we find that model attention consistently concentrates on the specific anatomical structures clinically relevant to each target finding (\textbf{Figure~\ref{fig:2}F}).
For example, for heart failure and ventricular dilation, \ourmethod focuses predominantly on the ventricular chambers and their boundaries, whereas for ventricular hypertrophy, attention shifts to the ventricular myocardium. These anatomically faithful attention patterns confirm that \ourmethod grounds its predictions in clinically meaningful visual evidence rather than spurious image artifacts.

\subsection{Video-report alignment for cross-modal diagnostic reasoning}

\begin{figure}[htbp]
  \centering
  \includegraphics[width=\linewidth]{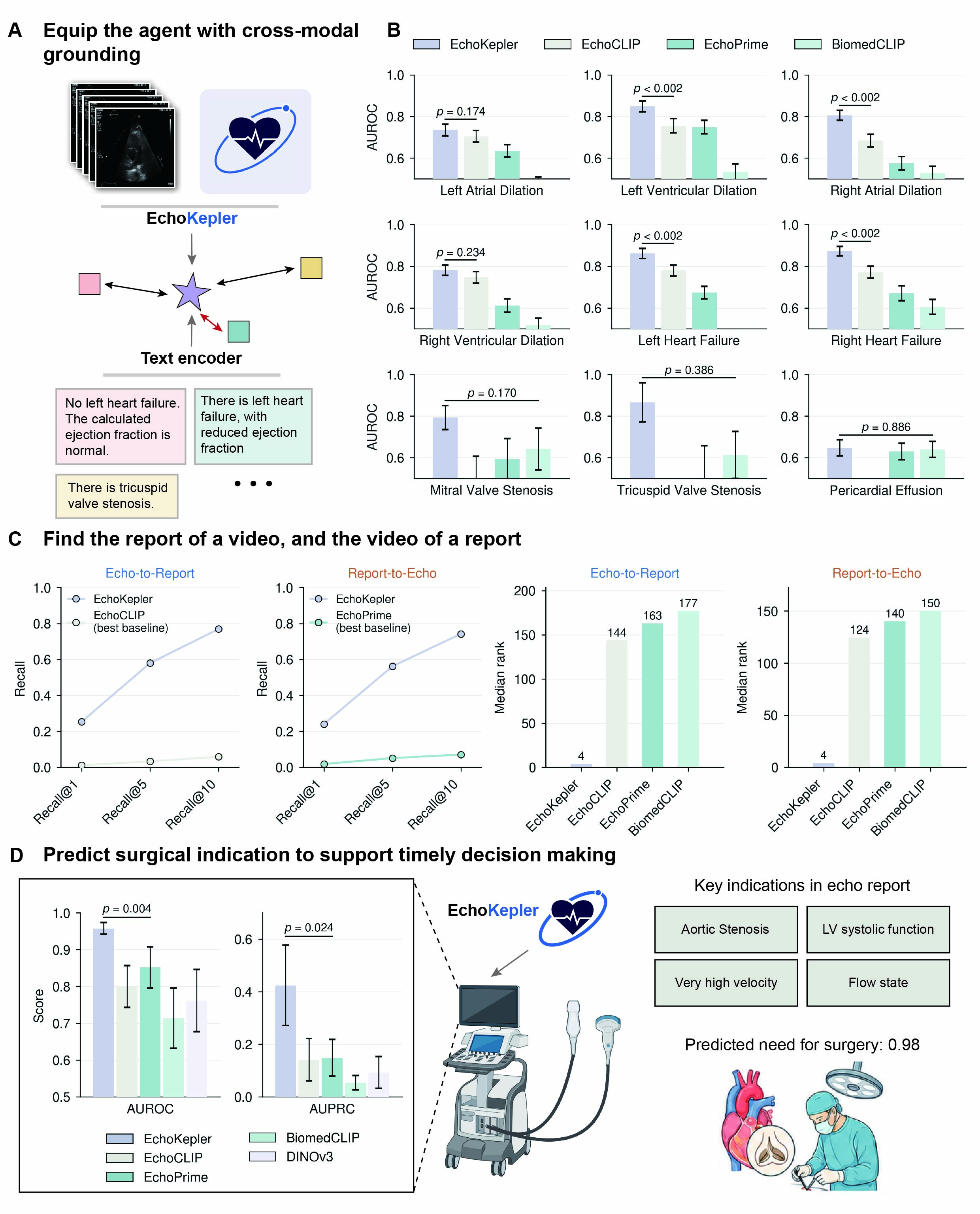}
  \caption{\textbf{Cross-modal evaluation of \ourmethod.} \textbf{A, } overview of zero-shot evaluation. \ourmethod converts echo videos in a study into one embedding and evidence queries into language embeddings. The prediction is made by finding the closest query in the latent space. \textbf{B, } bar plots comparing the zero-shot performance on detecting individual abnormalities. \textbf{C, } line charts indicating Recall$@k$ ($k = 1, 5, 10$) and bar plots comparing the median rank given a report to find a video or given a video to find a report. \textbf{D, } bar plots comparing \ourmethod and other methods in determining surgical indications from echocardiograms.}
  \label{fig:3}
\end{figure}

\begin{figure}[htbp]
  \centering
  \includegraphics[width=0.99\linewidth]{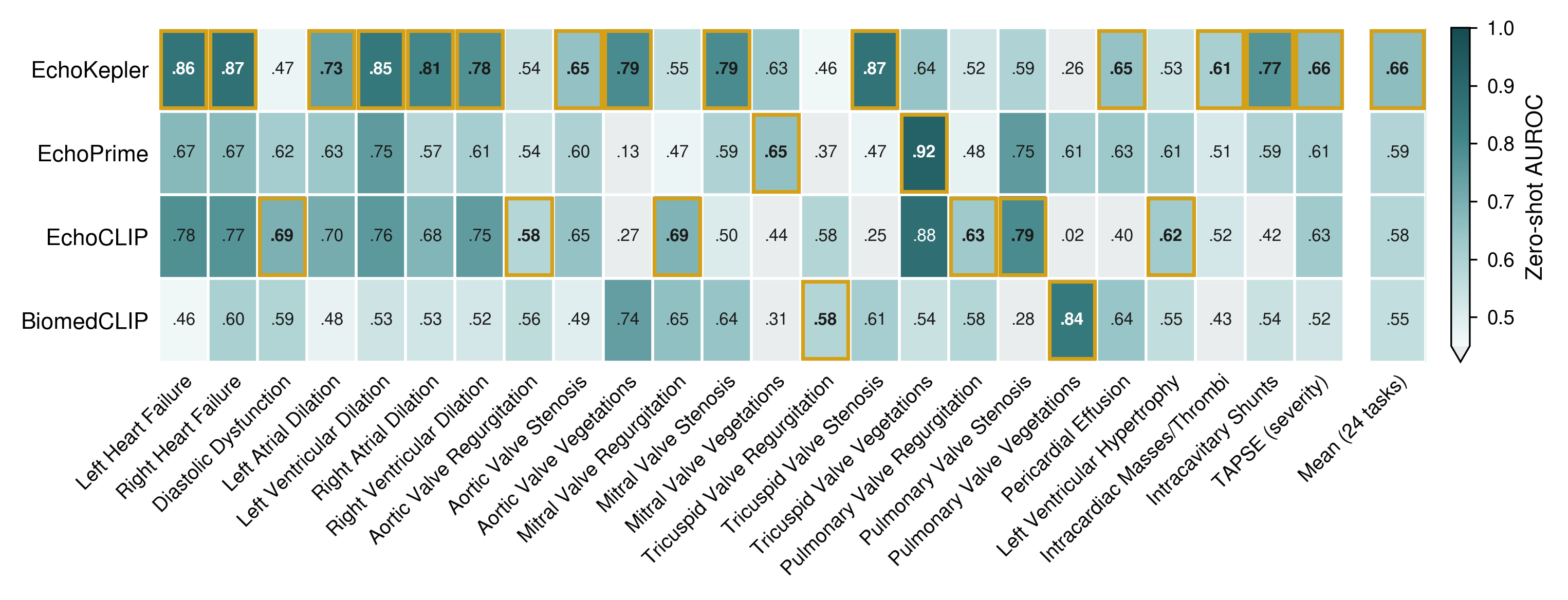}
  \caption{\textbf{Heatmap comparing the AUROC of \ourmethod and baselines for zero-shot abnormality detections.} Each row represents a model and each column represents an individual task. The gold box indicates the best model for each task.}
  \label{sfig:zeroshot_abnormality_auroc}
\end{figure}

\begin{figure}[htbp]
  \centering
  \includegraphics[width=0.99\linewidth]{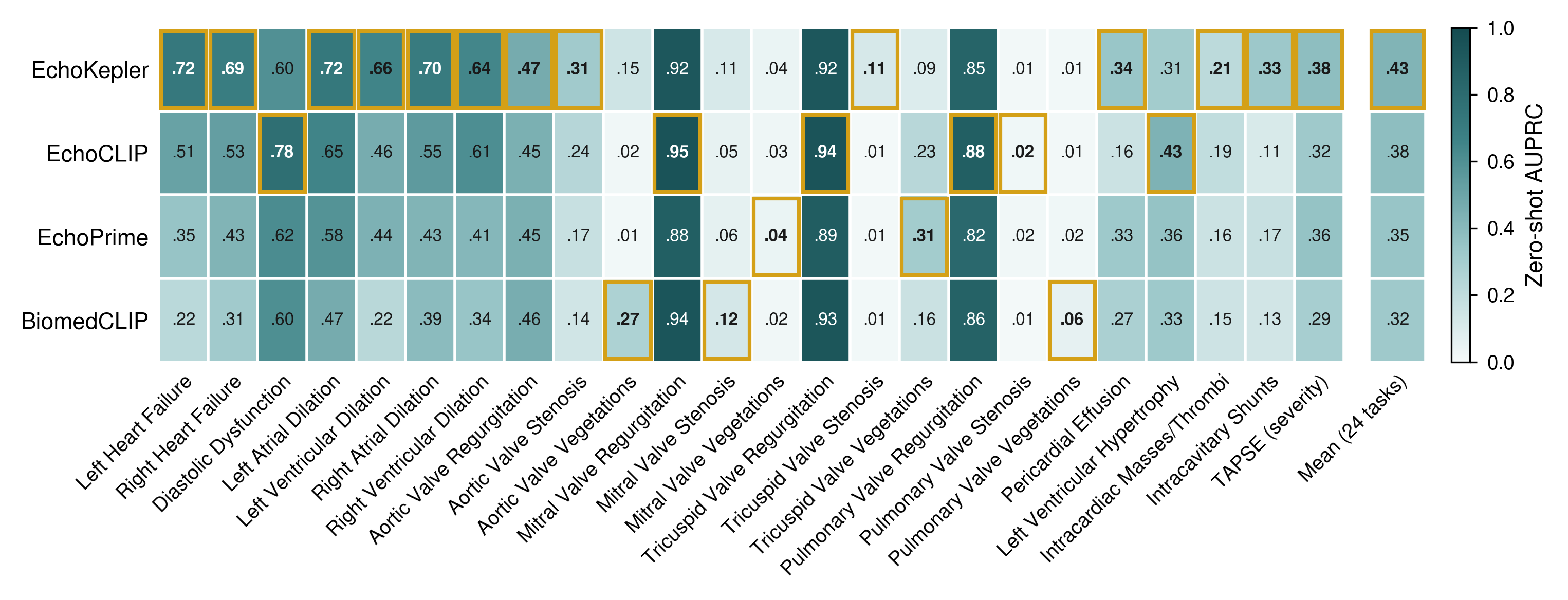}
  \caption{\textbf{Heatmap comparing the AUPRC of \ourmethod and baselines for zero-shot abnormality detections.} Each row represents a model and each column represents an individual task. The gold box indicates the best model for each task.}
  \label{sfig:zeroshot_abnormality_auprc}
\end{figure}

To align the vision encoder's embedding space with that of the LLM, we fine-tune it with a contrastive objective over paired clinical reports, enabling direct natural language querying for diagnostic evidence (\textbf{Figure~\ref{fig:3}A}). We first evaluate zero-shot detection across 24 abnormalities without task-specific supervised training. We benchmark \ourmethod against three vision-language baselines, including EchoCLIP, EchoPrime and BiomedCLIP, by prompting each model with clinically targeted textual queries and deriving corresponding prediction probabilities.
In \textbf{Figure~\ref{fig:3}B} and \textbf{Supplementary Figure~\ref{sfig:zeroshot_abnormality_auroc},~\ref{sfig:zeroshot_abnormality_auprc}}, we observe the superior zero-shot performance of \ourmethod for structural and functional findings across 24 tasks, with an 11.9\% improvement in AUROC. 
In detecting left and right heart failure, \ourmethod achieves zero-shot AUROCs of 0.86 and 0.87 respectively, whereas the best-performing baseline reached less than 0.8 in both. 
We then evaluate a more stringent test of alignment: the model must match a study to its own free-text report among all candidates. 
We conduct cross-modal retrieval across 395 held-out echocardiogram studies (\textbf{Figure~\ref{fig:3}C}).
\ourmethod retrieves the correct report for a given study with Recall@1 of 25.3\% and Recall@10 of 76.9\%, achieving a median rank of 4 out of 395. Comparable performance is observed in the reverse direction, from report to study. 
All three competing methods struggle in these tasks, with Recall@1 below 10\% and median ranks between 124 and 177. 
Together with the zero-shot results, these findings confirm that \ourmethod encodes echo studies in a space semantically aligned with clinical language, enabling open-vocabulary diagnostic reasoning beyond closed-set targets.

This strong performance further motivates us to examine whether the same representations support complex clinical decision-making. We therefore transition from predicting individual findings, such as aortic stenosis, LV systolic function, peak transvalvular velocity, and flow state, to determining whether their combination warrants surgical intervention. This presents a demanding yet consequential task, where reliably identifying surgical indications can assist clinicians in capturing the optimal therapeutic window.
We collect 5,524 patients from UW Medicine, of which 145 patients have surgical indications. The test patients remain strictly held out and untouched across all pretraining and downstream phases. 
In \textbf{Figure~\ref{fig:3}D}, \ourmethod demonstrate the superior performance over current foundation models, with an AUROC of 0.958 compared to 0.849 for the next best model. 
\ourmethod is the only method achieving a balanced accuracy above 80\%. 
Overall, the cross-modal alignment and clinical decision-making capabilities of our model demonstrate its potential utility in clinical practice.

\subsection{An agentic framework for selective diagnostic tool invocation}

\begin{figure}[htbp]
  \vspace*{-0.8cm}
  \centering
  \includegraphics[width=0.98\linewidth]{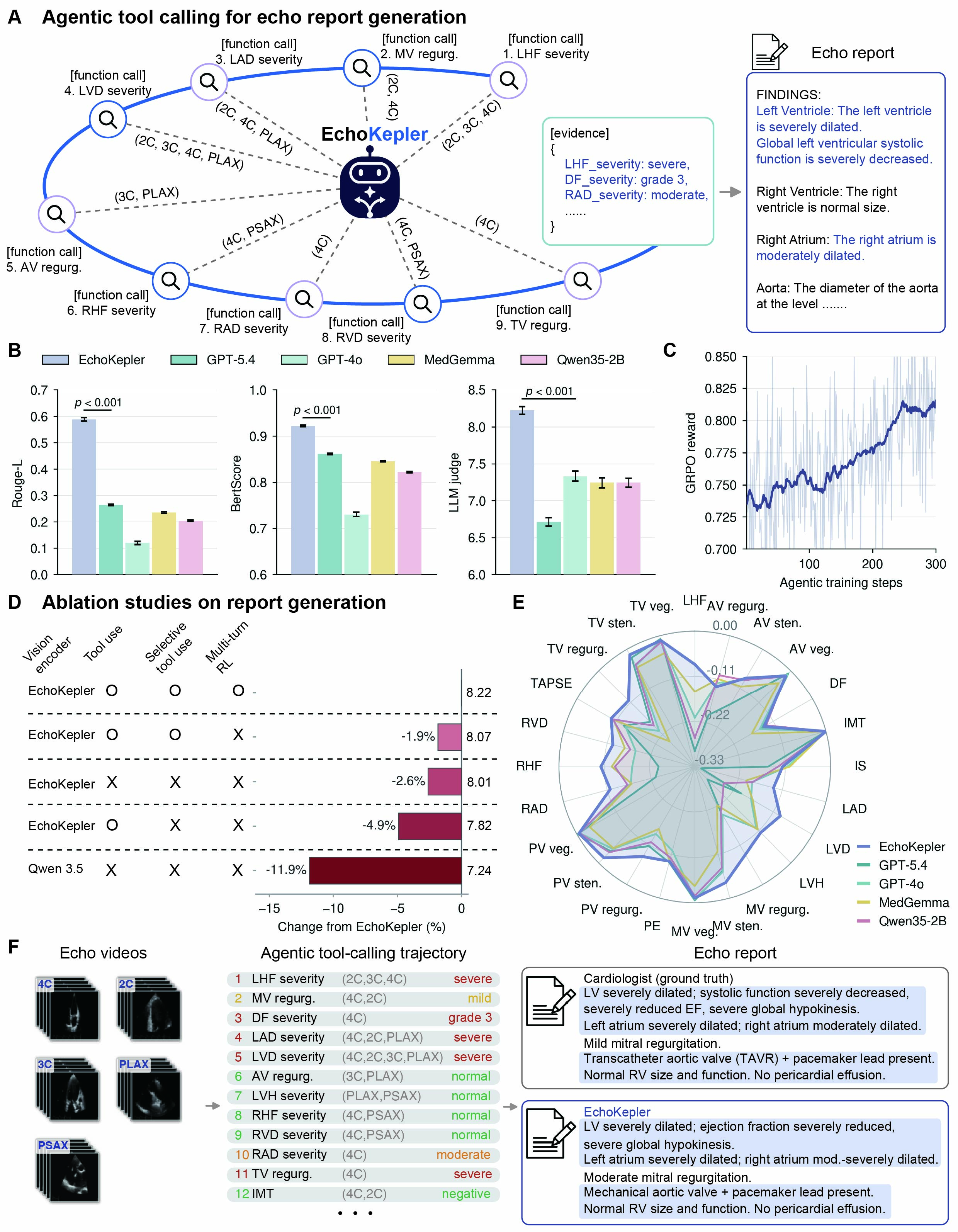}
  \vspace*{-0.3cm}
  \caption{\textbf{\ourmethod generates echo report through agentic tool calling.} \textbf{A, } Overview of the agentic framework. EchoKepler issues function calls to assess nine cardiac findings, selecting relevant echo views (2C/3C/4C/PLAX/PSAX) for each, then synthesizes the aggregated evidence into a structured echo report. \textbf{B, } Report-quality comparison against baselines on Rouge-L, BertScore, and the LLM-as-a-judge score. \textbf{C, } GRPO reward over agentic training steps. \textbf{D, } Ablation of key components (vision encoder, tool use, selective tool use, multi-turn RL); bars show the LLM-as-a-judge score change relative to the full model. \textbf{E, } Radar plot of per-condition performance across cardiac abnormalities for EchoKepler versus other methods. \textbf{F, } Case study: multi-view echo videos processed through a 14-step tool-calling trajectory.}
  \label{fig:4}
\end{figure}

The tool suite developed above enables step-by-step evidence collection for echocardiographic reporting, with one tool invocation at a time. However, patients exhibit substantial heterogeneity. A diagnostic tool gathering critical evidence for one case may be redundant or misleading for another. 
Therefore, the choice and execution schedule of diagnostic tools must be decided on a per-patient basis instead of a predetermined pipeline.
For example, in a patient with reduced left ventricular systolic function, quantifying mitral regurgitation and left atrial size helps establish the downstream consequences, whereas in a patient with suspected aortic stenosis, the informative evidence is instead the peak transvalvular velocity and left ventricular wall thickness. 
We train an agentic system to learn personalized and optimal tool invocation orders using a two-stage curriculum learning paradigm. 
First, we perform supervised fine-tuning over full tool calling trajectories synthesized by GPT-5.4. This stage equips the agent with structural schema and invocation syntax for orchestrating multi-turn tool invocations into diagnostic reports.
To incentivize optimal trajectory exploration, we then train the agent end-to-end via reinforcement learning using GRPO~\cite{liu2024deepseek}. Specifically, the alignment between the generated report and ground truth report serves as the reward for each sampled trajectory.
We observe that the training reward steadily increases across optimization steps, confirming that the agent effectively explores and converges toward optimal tool-calling trajectories (\textbf{Figure~\ref{fig:4}C}).
In inference, the \ourmethod agent first calls the view-classification tool to determine the acquisition view of each echo video. Then it invokes diagnostic tools to gather evidence one step at a time. 
At each turn, based on the initial observation and evidence collected so far, the agent chooses the tool to invoke and relevant views to supply as inputs. The returned diagnostic prediction is integrated into the accumulated evidence to inform future turns. Once the agent determines the evidence is sufficiently informative, it aggregates the findings into a JSON schema, prompting the system to generate the final clinical report (\textbf{Figure~\ref{fig:4}A}).

To evaluate performance, we compare against leading general-purpose frontier models, including GPT-5.4, GPT-4o and Qwen-3.5~\cite{teamqwen3}, which demonstrate strong clinical competence across healthcare benchmarks.
Additionally, we evaluate MedGemma v1.5 as a domain-specific multimodal baseline optimized for medical imaging and clinical reasoning.
Nevertheless, in \textbf{Figure~\ref{fig:4}B}, we observe \ourmethod substantially outperforms other models in lexical overlap (Rouge-L~\cite{lin2004rouge}), semantic similarity (BertScore~\cite{zhang2019bertscore}) and factual accuracy (LLM judge), indicating the critical role of multi-turn evidence collection through tool invocation. We also evaluate per-category factual accuracy across the 24 abnormalities using an LLM-as-a-judge evaluation, demonstrating that \ourmethod consistently outperforms competing models across all categories (\textbf{Figure~\ref{fig:4}E}).

To quantify how each design component contributes to performance, we conduct a series of ablation studies. Compared to Qwen-3.5, the vision encoder of \ourmethod yields a substantial gain, raising the LLM-as-a-judge score by 10.6\%. 
Naively invoking the entire tool set does not yield an improved score, highlighting the need for customized tool choices and execution order on a per-patient basis. In contrast, selective tool invocation shows superiority, delivering a 3.2\% increase in the overall score.
Subsequent reinforcement learning further increases the score from 8.07 to 8.22, enabling the agent to discover optimal tool-use strategies.

Finally, we examine an end-to-end case study illustrating how \ourmethod coordinates multi-turn tool invocations to generate high-fidelity reports. In a representative case characterized by severe dilated cardiomyopathy, an aortic valve prosthesis, and a pacemaker lead, the agent orchestrates fourteen tool calls across five available views (4C, 2C, 3C, PLAX, PSAX). Critically, it restricts tool inputs strictly to views with diagnostic utility for each specific abnormality. 
For example, it queries left-sided failure and chamber dimensions on apical and long-axis views, while isolating right-heart metrics and wall hypertrophy to views that resolve the corresponding anatomical structures.
The aggregated evidence forms a grounded clinical summary that drives high-quality report generation.

\subsection{Self-reflective evidence verification for longitudinal analysis}

\begin{figure}[htbp]
  \vspace*{-0.8cm}
  \centering
  \includegraphics[width=\linewidth]{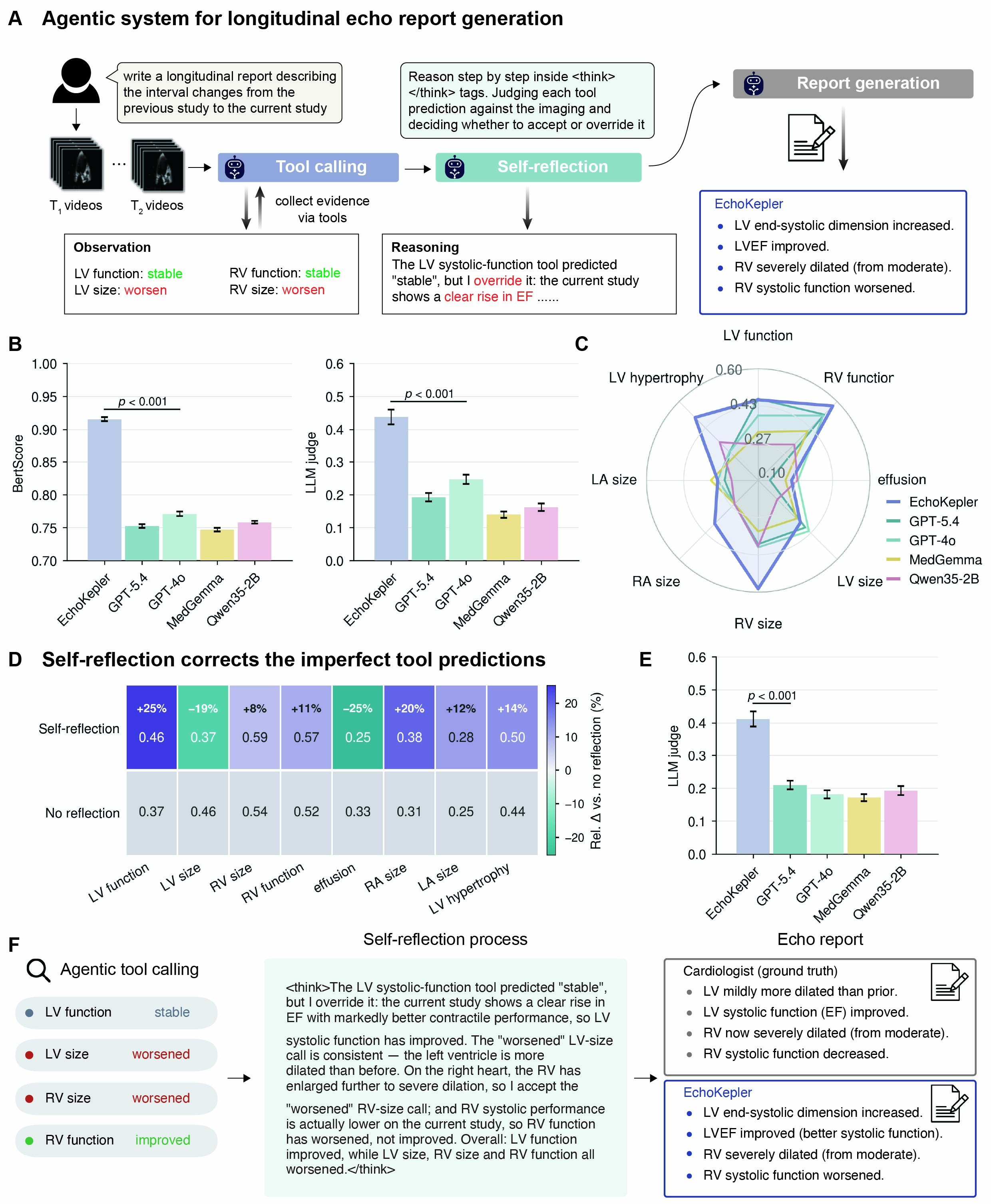}
  \vspace*{-0.3cm}
  \caption{\textbf{\ourmethod enables longitudinal analysis.} \textbf{A, } Overview of the agentic framework for longitudinal report generation with a self-reflection agent. \textbf{B, } Bar plots comparing \ourmethod and baselines in longitudinal report generation using BertScore and the LLM-as-a-judge score. \textbf{C, } Radar plot comparing the LLM-as-a-judge score across eight dimensions of interval changes. \textbf{D, } Ablation study demonstrating the performance gain from introducing the self-reflection agent. \textbf{E, } Bar plots comparing \ourmethod and baselines in forecasting future interval changes from current studies. \textbf{F, } A case study showing the agentic tool calling trace and self-reflection process used to generate a faithful longitudinal comparison report.}
  \label{fig:5}
\end{figure}

Longitudinal evaluations are essential for monitoring disease progression and determining the optimal window for clinical intervention. For example, valve replacement in asymptomatic severe valvular disease is often indicated by progressive decline across serial examinations. This setting is particularly challenging for AI systems, as minor errors in interpreting an individual study propagate across comparisons, leading to distorted estimates of interval change.
In a preliminary experiment, detecting interval changes in left atrial dilation (LAD) severity achieves an AUROC of only 0.723, substantially lower than the 0.833 attained by the corresponding single-study assessment tool.
Directly summarizing such unverified predictions into a report therefore risks propagating erroneous evidence into the final comparative assessment.

To rectify these errors, we introduce a self-reflection step to explicitly verify whether each tool prediction remains consistent with the observed imaging data (\textbf{Figure~\ref{fig:5}A}). Powered by the \ourmethod vision encoder, the agent generates a reasoning trace that grounds each detected mismatch in visual evidence, suggests a targeted revision, and subsequently compiles the verified findings into a longitudinal report. The entire agent is optimized end-to-end with reinforcement learning, using the diagnostic accuracy of the generated report as the reward signal to refine its verification strategy.

To evaluate \ourmethod in this challenging setting, we collect 63 held-out longitudinal pairs from UW Medicine, each comprising a prior and current study from the same patient. The task is to generate a longitudinal report detailing interval disease progression. We compare our approach against GPT-5.4, GPT-4o, MedGemma, and Qwen-3.5, each prompted with both examinations. While all models exhibit performance drops relative to single-study evaluations, \ourmethod outperforms competing baselines by a large margin. Specifically, it delivers comparable or markedly superior accuracy across eight dimensions of interval changes, including left and right atrial sizes, biventricular dimensions, left and right systolic function, pericardial effusion, and left ventricular hypertrophy (\textbf{Figure~\ref{fig:5}C}). Ablation experiments removing the self-reflection module reveal that self-reflective verification alone attains a 5.8\% gain in LLM-as-a-judge score, validating its necessity for robust comparative reasoning.

These promising results motivate us to explore whether the system can further forecast future interval progression when provided only with the current examination. As shown in \textbf{Figure~\ref{fig:5}E}, the superiority of \ourmethod persists in this challenging task, significantly outperforming the next strongest frontier baseline in LLM-as-a-judge score (0.41 versus 0.21). Notably, frontier models tend to hallucinate structural findings unsupported by evidence, whereas \ourmethod remains grounded in verified observations. Although preliminary, these gains highlight the framework's capacity to expand from retrospective diagnosis to clinical prognosis. In \textbf{Figure~\ref{fig:5}F}, we present a case study showcasing the agent's tool-calling trajectory and self-reflection mechanism during longitudinal report generation.

\section{Discussion}
\label{sec:discussion}
We present \ourmethod, an agentic framework capable of interpreting over a hundred videos within a single echocardiographic study to produce evidence-grounded clinical reports. The system operates across four coordinated stages: view prediction to identify acquisition views, multi-turn tool invocation to gather diagnostic evidence selectively across relevant views, self-reflection to verify and rectify findings, and report generation to finalize conclusions. The specialized diagnostic tools cover key clinical dimensions evaluated by cardiologists, including valvular diseases, chamber dilation, and ventricular function. Both the tool suite and the LLM agent are grounded in a domain-specific echocardiography foundation model pretrained on 30 million frames. Together, these components empower \ourmethod to mirror clinical reasoning. It can iteratively acquire evidence, navigate across multi-view data, and deliver fully traceable, hallucination-resistant clinical reports.

In our extensive evaluations, \ourmethod demonstrated substantial improvements over EchoCLIP, the strongest competing model, by 9.6\% across 24 abnormality detection tasks. For clinically consequential decision-making, \ourmethod achieved superior performance in identifying surgical indications with an AUROC of 0.958, highlighting its real-world utility. Furthermore, compared to existing echocardiography foundation models, \ourmethod is uniquely capable of generating accurate clinical reports, outperforming both general-purpose and specialized medical frontier models in an end-to-end evaluation. We attribute the success of \ourmethod to its selective tool invocations on a per-patient basis. This strategy not only improves diagnostic accuracy, but also grounds report statements in concrete evidence collected by the tools, thereby guarding against model hallucinations. This formulation captures the fundamental principle of agentic systems by decomposing a complex task into specialized sub-steps, each solved by a dedicated component before being composed into the final solution.

Whereas prior foundation models focus on learning transferable representations for various downstream classification tasks~\cite{xu2024whole, xu2025cardiac, liu2026three, usuyama2026gigapath}, \ourmethod advances this paradigm by building an agentic system on top of these representations. Consequently, our framework can tackle substantially more complex tasks, such as end-to-end report generation and longitudinal interpretation. In this work, we demonstrate two complementary ways to integrate a foundation model into an agentic system. First, it can be exposed as callable diagnostic tools that the agent selectively invokes. Second, it can serve as the vision encoder of the multimodal language model, ensuring that every reasoning step is grounded in domain-specialized visual representations.

Despite its strong performance, our study has several key limitations. First, \ourmethod was developed and validated on patient cohorts solely from UW Medicine. Although we hold out data across a broad range of clinical indications, external and prospective validation across diverse sites and populations is still needed to demonstrate robust generalizability. We also plan to open-source our model weights so that the community can evaluate on their own datasets, providing critical feedback for future improvements. Second, although the diagnostic tool suite covers common examinations, its predefined set of abnormalities may not accommodate unseen findings or measurements. A promising future direction is to adopt a self-evolving design~\cite{gao2025survey}, enabling the agent to autonomously collect data and train new diagnostic tools when encountering findings beyond its current scope. Third, longitudinal interpretation remains inherently challenging, as the interval-change prediction tools achieve lower performance than their single-study counterparts. Although the self-reflection agent partially mitigates these errors, future work is needed to incorporate larger datasets with richer supervision for serial comparisons.
\newpage
\section{Methods}
\label{sec:methods}
\subsection{Details of our data}
The UW Medicine cohort consists of 645,345 echo videos from 9,497 studies acquired between 2017 and 2018. The cohort includes transthoracic echocardiograms, congenital studies, stress tests, and transesophageal echocardiograms, reflecting an unselected, real-world spectrum of clinical indications. On average, each echocardiogram study consists of 68 videos acquired from multiple standard views, spanning parasternal (PLAX, PSAX), apical (2-, 3-, 4-, and 5-chamber), subcostal and vascular (IVC, abdominal aorta, hepatic vein) windows, as well as 3D acquisitions. Among these, 7,860 studies are paired with a free-text clinical report. We use these study-text paired data to supervise the vision-language alignment. We compile a comprehensive list of 24 abnormalities that a cardiologist routinely assesses. For each abnormality, the ground-truth label, including its presence and its severity, is extracted from the corresponding clinical report using an LLM. The extracted labels are used to train the suite of diagnostic tools.

For pretraining the foundation model in \ourmethod, we curate 30,080,731 2D echo frames, each resized to 256$\times$256 pixels. Raw studies are exported in DICOM format from the clinical system and deidentified by stripping the DICOM headers that have patient metadata. Every video is temporally normalized to 20 fps to enforce a consistent time step across acquisitions. All videos are then padded or cropped to 64 frames. Longer videos are center-cropped, symmetrically discarding frames from both ends. To pad shorter videos to the same length, we cyclically repeat their clips. Each frame is resized to a fixed 256 by 256 resolution using bilinear interpolation, and stored as three replicated grayscale channels. The data are split into training, validation, and test sets at a 90:5:5 ratio at the study level prior to any training, and the test set is held out untouched throughout every stage of model training. For vision-language alignment, we de-identify echocardiography reports by removing patient metadata and retaining primarily the findings and conclusions sections.

All downstream tasks are derived from the subset of studies paired with free-text clinical reports. To prevent data leakage, these downstream tasks inherit the same study-level partition described above. For abnormality detection, all 24 abnormalities are labeled as present or absent. We also have severity grades available for 18 of them. The dataset size varies across abnormalities depending on how frequently each finding is documented. The training and test sample distributions, availability of severity grading, and full names and abbreviations of all abnormalities are provided in \textbf{Supplementary Table~\ref{tab:supp_abnormalities}}. For report generation, we construct both the findings and conclusions sections as generation targets. To make sure generation targets are answerable from imaging alone instead of requiring other quantitative measurements, these targets are rewritten using a large language model to remove numeric measurements while preserving all qualitative descriptions.

We also label whether a study documents indications for aortic valve surgery in aortic stenosis. Labels are extracted from echo reports using a large language model prompted with guideline criteria that define this situation (\textbf{Supplementary Figure~\ref{sfig:supp_surgery_prompt}}). There are three conditions which indicate surgery: (1) symptomatic severe aortic stenosis; (2) severe stenosis with low flow, low gradient and preserved ejection fraction; and (3) severe stenosis with low flow, low gradient and left ventricular systolic dysfunction. We prompt the large language model to return the verbatim supporting texts, a three-way decision, a confidence score, and the evidence it relies on. To check extraction quality, we ask an experienced clinician who has more than ten years' practice in the field to manually review 10 randomly sampled studies with 4 positive, 5 negative, and 1 unknown case. The clinician agrees with every extracted case.

In this cohort, 1,138 patients undergo more than one echocardiogram study, and each study has its own clinical report. 770 of them are examined twice and the others even more (up to fourteen times). We construct 1,838 longitudinal pairs taking the temporally consecutive studies, with a median interval of 11 days and an interquartile range of 2 to 52 days. For each pair, we prompt a large language model to compare the two single-study reports and describe the interval change (\textbf{Supplementary Figure~\ref{sfig:supp_longitudinal_prompt}}). To assess the quality of the generation, we do a spot check by asking experienced clinicians to review 50 results against the pairs of source reports they are derived from. The clinicians confirm the synthesized reports faithfully describe the interval changes in the documented abnormalities.

\subsection{Details of pretraining the echo foundation model}
We conduct a two-stage curriculum learning to pretrain the echo foundation model. In the first stage, we pretrain the foundation model to learn frame-level representations. In the second stage, we train an additional aggregator to learn study-level representations using study-report pairs with a contrastive alignment loss.

We employ DINOv3 self-supervised learning in the first stage. Our pretraining corpus includes 30,080,731 frames extracted from echo videos at their native acquisition FPS. The vision encoder is a large vision transformer model (ViT-L) with 300M parameters with a patch size of 16, initialized from the public DINOv3 checkpoint pretrained on LVD-1689M. In the multi-cropping process, two global crops are sampled at scales from 0.32 to 1 of the images, resized to $224 \times 224$ pixels. The local crops are sampled at scales from 0.05 to 0.32, resized to $96 \times 96$ pixels. Each frame is normalized using the standard ImageNet mean and standard deviation. We use the default 262,144 prototypes in self-distillation and the Sinkhorn-Knopp centering strategy. We optimize the model with the combination of DINO loss, iBOT loss, and KoLeo regularizer for 500,000 iterations on 4$\times$80GB A100 GPU devices, with a global batch size of 192. Subsequently, we continue from the resulting teacher checkpoint for a further 500,000 iterations of training with the addition of Gram anchoring loss, whose weight is annealed from 0.0 to 0.2.

After learning the frame-level vision encoder, we build an attention-based, multiple-instance learning head that aggregates the frames of each video into a video-level embedding and averages these into a single study-level representation. To make this representation clinically meaningful, we align it with the embedding of the corresponding clinical report, obtained from a domain-specific text encoder, PubMedBERT~\cite{gu2021domain}. We use the CLIP~\cite{radford2021learning} loss as the contrastive alignment objective. The ViT-L encoder, the ABMIL~\cite{ilse2018attention} head, and the text encoder are jointly optimized for 30 epochs with a global batch size of 80 studies. In each training iteration, we randomly sample 4 videos from a single study. At inference, a study-level embedding is obtained by encoding all videos of the study; a view-informed embedding is created by encoding only the videos acquired in a specified view.

\subsection{Details of developing a suite of diagnostic tools}
To give \ourmethod tools to collect evidence for echocardiographic interpretation, we train a set of 24 classifiers on top of the view-informed embeddings produced by the \ourmethod foundation model. Each classifier is a single linear layer mapping an embedding to class logits of an abnormality in \textbf{Supplementary Table~\ref{tab:supp_abnormalities}}. When the four-level severity grades are required, we supervise the classifier with severity rather than presence, so that the tool reports evidence at the finer granularity. Each classifier is trained for 100 epochs on a single A100 GPU with a batch size of 512 and a base learning rate of $5\times10^{-4}$, and we oversample the minority classes so that every class is represented equally during training. The best checkpoint is selected on validation AUROC. Each trained classifier is then exposed to the agent as a function. It takes the videos of the views relevant to its target abnormality and returns the predicted severity, or presence and absence. To determine the associated views for abnormalities, we use a static map verified by both the medical literature and experienced clinicians (\textbf{Supplementary Table~\ref{tab:supp_task_views}}).

In comparison with other competing methods on abnormality detection (\textbf{Figure~\ref{fig:2}B}), we focus on evaluating the representation quality. All encoders are evaluated under the identical linear probing protocol. We compare our echo foundation encoder against EchoCLIP, EchoPrime, BiomedCLIP, and publicly available DINOv3 encoders. EchoCLIP, BiomedCLIP and DINOv3 encode frames independently and average them into a video embedding. EchoPrime is a video transformer (mViT-v2-S) that consumes 16 frames jointly and produces a video embedding directly. The embedding at the study-level is the mean of all videos in the study. Each model is processed through the same pipeline. Each linear probe is trained for 100 epochs with a batch size of 512, a base learning rate of 5e-4, and class-balanced sampling.

\subsection{Details of training the agentic system}
In the agentic system, the view agent is a hard-coded program which automatically calls the view classification tool to determine how each video in a study is acquired. The view classification tool is trained on expert manual annotations covering 2,050 videos from 34 studies spanning 10 standard views, including PLAX, PSAX, 2C, 3C, 4C, 5C, IVC, focus on abdominal aorta, focus on hepatic vein and 3D echo. This tool is instantiated as a linear classifier trained on a 1024-dimensional video embedding from the \ourmethod foundation model. It achieves an AUROC of 0.963 on a held-out test set, indicating that the acquisition view information is highly predictable from individual videos. The predicted view information is used to gather relevant videos as inputs to diagnostic tools via the static map in \textbf{Supplementary Table~\ref{tab:supp_task_views}}.

The multi-turn tool-calling agent, the self-reflection agent, and the report agent are all powered by a single vision-language model (VLM) adapted to echocardiography interpretation. We build our model on Qwen2.5-VL-3B-Instruct and replace its general-purpose vision tower with our echo foundation encoder, a ViT-L pretrained on 30,080,731 echocardiographic frames. Frame-level features are pooled into a 1024-dimensional video embedding with ABMIL. Therefore, a study with an arbitrary number of videos is input to the VLM as a sequence of per-video tokens. Because the direct substitution might break the vision-language alignment that Qwen2.5-VL acquires during its pretraining, the replacement projector is randomly initialized, and the language model is optimized to understand the echo video embeddings. We therefore jointly optimize the projector and the language model using full-parameter supervised finetuning.

We construct a large-scale instruction corpus of 251,661 samples drawn from 6,679 studies and spanning five question-answering formats, grouped into two categories: (i) view-grounded single-task assessment, where the model answers one of 24 abnormality or severity questions either directly from the assigned views or by first invoking a tool call and then interpreting the returned result; (ii) report generation, including direct video-to-report generation, evidence-to-report synthesis from structured findings, and tool-augmented report generation in which the model gathers evidence in a multi-turn tool calling scheme. The multi-turn tool-calling trajectories are synthesized by GPT-5.4 rollout against a ground-truth environment. For each study, we expose only the tools whose target abnormalities have been examined by cardiologists and have clinical labels in the ground truth report. Then we let GPT choose one tool per turn, and return the clinical label as the observation, so that at each step the model is conditioned on gathered evidence. We first perform the full-parameter supervised finetuning for 25,101 optimizer steps (3 epochs) with a peak learning rate of 1.0e-5, a cosine decay schedule with 3\% linear warmup, and an effective batch size of 32 (per-device batch of 2 with a gradient accumulation of 4 across 4 GPUs). During training, we use bf16 precision, DeepSpeed ZeRO-2 sharding, and gradient checkpointing, with a maximum sequence length of 16,384 tokens. The training phase takes approximately 72 GPU-hours per device.

After supervised finetuning, we further improve the agent's ability to find an optimal tool-calling trajectory for report generation with reinforcement learning (RL). The RL phase teaches the model to act in the real-world environment. During rollout time, each tool call is executed using the deployed diagnostic tools, linear classifiers built upon the pre-computed video embeddings by the pretrained foundation model. After a tool is executed and its prediction is returned as an observation, the agent decides which tool to call next until it gathers enough evidence. We use Group Relative Policy Optimization (GRPO) to train the multi-turn tool-calling agent, with BLEU-4 between the generated and reference reports serving as the reward signal.
We initialize the policy from the SFT checkpoint, keeping the embedding projector frozen so that the visual interface established during SFT is preserved. We use a learning rate of 1e-6 for GRPO, a batch size of 16 per step, a low-variance KL penalty with coefficient 0.001 for 3 epochs on 4 A100-80GB GPUs. We generate eight rollouts per input using SGLang~\cite{zheng2024sglang} (\texttt{rollout.n}=8).
Similarly, we use GRPO to optimize the reasoning process inside the  tokens of the self-reflection agent. We synthesize the reasoning traces with GPT-5.4, prompted to imitate an expert verifying the tool-returned evidence step by step and explicitly overriding any tool prediction that contradicts the imaging. We then apply GRPO with a learning rate of 1e-6, a batch size of 16 per step and eight rollouts per input, a low-variance KL penalty with coefficient 0.001, and a response budget of 2,048 tokens, for 5 epochs on the same 4 A100-80GB configuration. The reward is an LLM-as-judge score (GPT-4o) computed on the report emitted after the final \texttt{</think>}. The reasoning trace itself is never scored directly, so the model is free to discover whatever verification strategy yields a more accurate report.

\subsection{Details of longitudinal report generation}
To generate longitudinal reports, we use the same agentic system as for the single-study report. The only difference is that we replace the suite of diagnostic tools with eight interval-change prediction models, covering left ventricular systolic function, left ventricular size, left ventricular hypertrophy, left atrial size, right ventricular size and function, right atrial size, and right heart function. Each change model uses the frozen study embeddings of the two visits. A single linear layer is trained on the concatenated embeddings [prev, cur, cur - prev] (3 $\times$ 1024 = 3072 dimensions), predicting one of three interval-change classes (improved / stable / worsened). Pooling follows the same view-grounded protocol as the single-study tools, so each dimension is computed only from the views in which it is assessable. The view agent, the multi-turn tool-calling agent, the self-reflection agent, and the report agent remain unchanged. The agentic framework first calls interval-change prediction tools, then verifies and corrects potential errors before consolidating the evidence into a longitudinal report. For the forecasting experiments, we follow the same pipeline, except that both the agents and the prediction tools observe a single baseline study and forecast the interval changes expected at the next follow-up. The eight change models are accordingly re-trained on the baseline study embeddings alone. The generated report describes what is expected to change rather than what has already changed.

\subsection{Evaluation and benchmarking}
We evaluate the generated echocardiography reports along three complementary axes. We report ROUGE-L for lexical overlap, BERTScore for semantic similarity, and an LLM-as-judge score for factual accuracy.

We compute ROUGE-L with the standard rouge\_score implementation and Porter stemming and report the F1 of the longest common subsequence between the generated and reference reports. We compute the BERTScore with a DistilBERT~\cite{sanh2019distilbert} encoder and report the F1 of the greedily matched token embeddings. Both metrics reward surface agreement, but neither can tell a correct finding from its negation. We therefore add a third metric that scores clinical content directly, i.e. LLM-as-judge.

For this method's score, we prompt GPT-4o with the reference report, the generated report, and a list of 24 diagnostic tasks (\textbf{Supplementary Figure~\ref{sfig:supp_judge_report_prompt}}). The judge compares the two reports task by task and states what each one says about that task. It then assigns a deduction of 0 when the two agree, 0.1 when the prediction is slightly inaccurate, 0.2 when it differs noticeably, and 0.5 when it contradicts the reference. Hallucinated findings are penalized asymmetrically: if the reference is silent, an invented abnormal finding costs 0.5, whereas an invented normal or mild finding costs 0.1. The final score is $\max(0, 10 - \sum \text{deductions})$, so a report that agrees with the reference on every task scores 10. We report the mean score over the test set.

We use a different setup for LLM-as-judge for the longitudinal reports, because a longitudinal report asserts a direction of change rather than an absolute finding. We therefore score each report per cardiac dimension. We prompt GPT-5.4 with the reference and the generated report, and for every dimension the reference mentions, the judge emits a label describing how the prediction behaves (\textbf{Supplementary Figure~\ref{sfig:supp_judge_longitudinal_prompt}}). We define the LLM-as-judge scores as: 1.00 when both direction and magnitude are correct, 0.75 when the direction is correct but the magnitude is off by one grade, 0.50 when the direction is correct but the magnitude is off by two or more grades, 0.25 when one report claims a change and the other calls it stable, 0.25 when the prediction does not address the dimension at all, and 0.00 when the prediction asserts the opposite direction.

\section*{Data Availability}
The echocardiography studies and their paired clinical reports were collected at the University of Washington under an approved institutional review board protocol and contain protected health information. They therefore cannot be released publicly. De-identified data may be made available to qualified researchers for non-commercial use upon reasonable request to the corresponding author, subject to institutional approval and a data use agreement. The publicly available models used as baselines, EchoCLIP, EchoPrime, BiomedCLIP, DINOv2 and DINOv3, remain available from their original sources.

\section*{Code Availability}
All code for pretraining, agent training, reinforcement learning and evaluation will be available at \url{https://github.com/HanwenXuTHU/EchoKepler}. The repository includes the diagnostic tool environment, the data-construction scripts and the prompts reproduced in the Supplementary Figures. Trained model weights are released alongside the code.

\clearpage
\printbibliography

@article{papolos2016us,
  title={US hospital use of echocardiography: insights from the nationwide inpatient sample},
  author={Papolos, Alexander and Narula, Jagat and Bavishi, Chirag and Chaudhry, Farooq A and Sengupta, Partho P},
  journal={Journal of the American College of Cardiology},
  volume={67},
  number={5},
  pages={502--511},
  year={2016},
  publisher={American College of Cardiology Foundation Washington, DC}
}

@article{virnig2014trends,
  title={Trends in the Use of Echocardiography, 2007 to 2011},
  author={Virnig, Beth A and Shippee, Nathan D and O'Donnell, Brian and Zeglin, Jessica and Parashuram, Shriram},
  journal={Data Points Publication Series [Internet]},
  year={2014},
  publisher={Agency for Healthcare Research and Quality (US)}
}

@article{gu2021domain,
  title={Domain-specific language model pretraining for biomedical natural language processing},
  author={Gu, Yu and Tinn, Robert and Cheng, Hao and Lucas, Michael and Usuyama, Naoto and Liu, Xiaodong and Naumann, Tristan and Gao, Jianfeng and Poon, Hoifung},
  journal={ACM Transactions on Computing for Healthcare (HEALTH)},
  volume={3},
  number={1},
  pages={1--23},
  year={2021},
  publisher={ACM, New York, NY}
}

@inproceedings{ilse2018attention,
  title={Attention-based deep multiple instance learning},
  author={Ilse, Maximilian and Tomczak, Jakub and Welling, Max},
  booktitle={International conference on machine learning},
  pages={2127--2136},
  year={2018},
  organization={Pmlr}
}

@inproceedings{radford2021learning,
  title={Learning transferable visual models from natural language supervision},
  author={Radford, Alec and Kim, Jong Wook and Hallacy, Chris and Ramesh, Aditya and Goh, Gabriel and Agarwal, Sandhini and Sastry, Girish and Askell, Amanda and Mishkin, Pamela and Clark, Jack and others},
  booktitle={International conference on machine learning},
  pages={8748--8763},
  year={2021},
  organization={PmLR}
}

@article{sanh2019distilbert,
  title={DistilBERT, a distilled version of BERT: smaller, faster, cheaper and lighter},
  author={Sanh, Victor and Debut, Lysandre and Chaumond, Julien and Wolf, Thomas},
  journal={arXiv preprint arXiv:1910.01108},
  year={2019}
}

@article{zheng2024sglang,
  title={Sglang: Efficient execution of structured language model programs},
  author={Zheng, Lianmin and Yin, Liangsheng and Xie, Zhiqiang and Sun, Chuyue and Huang, Jeff and Yu, Cody H and Cao, Shiyi and Kozyrakis, Christos and Stoica, Ion and Gonzalez, Joseph E and others},
  journal={Advances in neural information processing systems},
  volume={37},
  pages={62557--62583},
  year={2024}
}

@article{colebourn2024state,
  title={State of the art: a roadmap for the national echocardiography team 2023},
  author={Colebourn, Claire},
  journal={Future Healthcare Journal},
  volume={11},
  number={1},
  pages={100129},
  year={2024},
  publisher={Elsevier}
}

@article{bonow1998guidelines,
  title={Guidelines for the management of patients with valvular heart disease: executive summary A report of the American College of Cardiology/American Heart Association Task Force on Practice Guidelines (Committee on Management of Patients with Valvular Heart Disease)},
  author={Bonow, Robert O and Carabello, Blase and de Leon Jr, Antonio C and Edmunds Jr, L Henry and Fedderly, Bradley J and Freed, Michael D and Gaasch, William H and McKay, Charles R and Nishimura, Rick A and O'Gara, Patrick T and others},
  journal={Circulation},
  volume={98},
  number={18},
  pages={1949--1984},
  year={1998},
  publisher={Lippincott Williams \& Wilkins}
}

@article{heidenreich20222022,
  title={2022 AHA/ACC/HFSA guideline for the management of heart failure: executive summary: a report of the American College of Cardiology/American Heart Association Joint Committee on Clinical Practice Guidelines},
  author={Heidenreich, Paul A and Bozkurt, Biykem and Aguilar, David and Allen, Larry A and Byun, Joni J and Colvin, Monica M and Deswal, Anita and Drazner, Mark H and Dunlay, Shannon M and Evers, Linda R and others},
  journal={Journal of the American College of Cardiology},
  volume={79},
  number={17},
  pages={1757--1780},
  year={2022},
  publisher={American College of Cardiology Foundation Washington DC}
}

@article{mitchell2019guidelines,
  title={Guidelines for performing a comprehensive transthoracic echocardiographic examination in adults: recommendations from the American Society of Echocardiography},
  author={Mitchell, Carol and Rahko, Peter S and Blauwet, Lori A and Canaday, Barry and Finstuen, Joshua A and Foster, Michael C and Horton, Kenneth and Ogunyankin, Kofo O and Palma, Richard A and Velazquez, Eric J},
  journal={Journal of the American Society of Echocardiography},
  volume={32},
  number={1},
  pages={1--64},
  year={2019},
  publisher={Elsevier}
}

@article{zhang2018fully,
  title={Fully automated echocardiogram interpretation in clinical practice: feasibility and diagnostic accuracy},
  author={Zhang, Jeffrey and Gajjala, Sravani and Agrawal, Pulkit and Tison, Geoffrey H and Hallock, Laura A and Beussink-Nelson, Lauren and Lassen, Mats H and Fan, Eugene and Aras, Mandar A and Jordan, ChaRandle and others},
  journal={Circulation},
  volume={138},
  number={16},
  pages={1623--1635},
  year={2018},
  publisher={Lippincott Williams \& Wilkins Hagerstown, MD}
}

@inproceedings{huang2023detecting,
  title={Detecting heart disease from multi-view ultrasound images via supervised attention multiple instance learning},
  author={Huang, Zhe and Wessler, Benjamin S and Hughes, Michael C},
  booktitle={Machine Learning for Healthcare Conference},
  pages={285--307},
  year={2023},
  organization={PMLR}
}

@article{vrudhula2024high,
  title={High-throughput deep learning detection of mitral regurgitation},
  author={Vrudhula, Amey and Duffy, Grant and Vukadinovic, Milos and Liang, David and Cheng, Susan and Ouyang, David},
  journal={Circulation},
  volume={150},
  number={12},
  pages={923--933},
  year={2024},
  publisher={Lippincott Williams \& Wilkins Hagerstown, MD}
}

@article{nagueh2016recommendations,
  title={Recommendations for the evaluation of left ventricular diastolic function by echocardiography: an update from the American Society of Echocardiography and the European Association of Cardiovascular Imaging},
  author={Nagueh, Sherif F and Smiseth, Otto A and Appleton, Christopher P and Byrd, Benjamin F and Dokainish, Hisham and Edvardsen, Thor and Flachskampf, Frank A and Gillebert, Thierry C and Klein, Allan L and Lancellotti, Patrizio and others},
  journal={Journal of the American Society of Echocardiography},
  volume={29},
  number={4},
  pages={277--314},
  year={2016},
  publisher={Elsevier}
}

@article{zoghbi2017recommendations,
  title={Recommendations for noninvasive evaluation of native valvular regurgitation: a report from the American Society of Echocardiography developed in collaboration with the Society for Cardiovascular Magnetic Resonance},
  author={Zoghbi, William A and Adams, David and Bonow, Robert O and Enriquez-Sarano, Maurice and Foster, Elyse and Grayburn, Paul A and Hahn, Rebecca T and Han, Yuchi and Hung, Judy and Lang, Roberto M and others},
  journal={Journal of the American Society of Echocardiography},
  volume={30},
  number={4},
  pages={303--371},
  year={2017},
  publisher={Elsevier}
}

@article{zambrano2025clinically,
  title={A clinically accessible small multimodal radiology model and evaluation metric for chest X-ray findings},
  author={Zambrano Chaves, Juan Manuel and Huang, Shih-Cheng and Xu, Yanbo and Xu, Hanwen and Usuyama, Naoto and Zhang, Sheng and Wang, Fei and Xie, Yujia and Khademi, Mahmoud and Yang, Ziyi and others},
  journal={Nature Communications},
  volume={16},
  number={1},
  pages={3108},
  year={2025},
  publisher={Nature Publishing Group UK London}
}

@article{li2023llava,
  title={Llava-med: Training a large language-and-vision assistant for biomedicine in one day},
  author={Li, Chunyuan and Wong, Cliff and Zhang, Sheng and Usuyama, Naoto and Liu, Haotian and Yang, Jianwei and Naumann, Tristan and Poon, Hoifung and Gao, Jianfeng},
  journal={Advances in Neural Information Processing Systems},
  volume={36},
  pages={28541--28564},
  year={2023}
}

@article{zhang2024generalist,
  title={A generalist vision--language foundation model for diverse biomedical tasks},
  author={Zhang, Kai and Zhou, Rong and Adhikarla, Eashan and Yan, Zhiling and Liu, Yixin and Yu, Jun and Liu, Zhengliang and Chen, Xun and Davison, Brian D and Ren, Hui and others},
  journal={Nature medicine},
  volume={30},
  number={11},
  pages={3129--3141},
  year={2024},
  publisher={Nature Publishing Group US New York}
}

@article{hamamci2026generalist,
  title={Generalist foundation models from a multimodal dataset for 3D computed tomography},
  author={Hamamci, Ibrahim Ethem and Er, Sezgin and Wang, Chenyu and Almas, Furkan and Simsek, Ayse Gulnihan and Esirgun, Sevval Nil and Dogan, Irem and Durugol, Omer Faruk and Hou, Benjamin and Shit, Suprosanna and others},
  journal={Nature Biomedical Engineering},
  pages={1--19},
  year={2026},
  publisher={Nature Publishing Group UK London}
}

@article{chao2025evaluating,
  title={Evaluating large language models in echocardiography reporting: opportunities and challenges},
  author={Chao, Chieh-Ju and Banerjee, Imon and Arsanjani, Reza and Ayoub, Chadi and Tseng, Andrew and Delbrouck, Jean-Benoit and Kane, Garvan C and Lopez-Jimenez, Francisco and Attia, Zachi and Oh, Jae K and others},
  journal={European Heart Journal-Digital Health},
  volume={6},
  number={3},
  pages={326--339},
  year={2025},
  publisher={Oxford University Press UK}
}

@article{tu2024towards,
  title={Towards generalist biomedical AI},
  author={Tu, Tao and Azizi, Shekoofeh and Driess, Danny and Schaekermann, Mike and Amin, Mohamed and Chang, Pi-Chuan and Carroll, Andrew and Lau, Charles and Tanno, Ryutaro and Ktena, Ira and others},
  journal={Nejm Ai},
  volume={1},
  number={3},
  pages={AIoa2300138},
  year={2024},
  publisher={Massachusetts Medical Society}
}

@article{christensen2024vision,
  title={Vision--language foundation model for echocardiogram interpretation},
  author={Christensen, Matthew and Vukadinovic, Milos and Yuan, Neal and Ouyang, David},
  journal={Nature Medicine},
  volume={30},
  number={5},
  pages={1481--1488},
  year={2024},
  publisher={Nature Publishing Group US New York}
}

@article{vukadinovic2026comprehensive,
  title={Comprehensive echocardiogram evaluation with view primed vision language AI},
  author={Vukadinovic, Milos and Chiu, I-Min and Tang, Xiu and Yuan, Neal and Chen, Tien-Yu and Cheng, Paul and Li, Debiao and Cheng, Susan and He, Bryan and Ouyang, David},
  journal={Nature},
  volume={650},
  number={8103},
  pages={970--977},
  year={2026},
  publisher={Nature Publishing Group UK London}
}

@article{jiang2025adaptation,
  title={Adaptation of agentic ai},
  author={Jiang, Pengcheng and Lin, Jiacheng and Shi, Zhiyi and Wang, Zifeng and He, Luxi and Wu, Yichen and Zhong, Ming and Song, Peiyang and Zhang, Qizheng and Wang, Heng and others},
  journal={arXiv preprint arXiv:2512.16301},
  year={2025}
}

@inproceedings{yang2024swe,
  title={Swe-agent: Agent-computer interfaces enable automated software engineering},
  author={Yang, John and Jimenez, Carlos E and Wettig, Alexander and Lieret, Kilian and Yao, Shunyu and Narasimhan, Karthik R and Press, Ofir},
  booktitle={The Thirty-eighth Annual Conference on Neural Information Processing Systems},
  year={2024}
}

@article{wang2024survey,
  title={A survey on large language model based autonomous agents},
  author={Wang, Lei and Ma, Chen and Feng, Xueyang and Zhang, Zeyu and Yang, Hao and Zhang, Jingsen and Chen, Zhiyuan and Tang, Jiakai and Chen, Xu and Lin, Yankai and others},
  journal={Frontiers of Computer Science},
  volume={18},
  number={6},
  pages={186345},
  year={2024},
  publisher={Springer}
}

@article{sun2025scaling,
  title={Scaling long-horizon llm agent via context-folding},
  author={Sun, Weiwei and Lu, Miao and Ling, Zhan and Liu, Kang and Yao, Xuesong and Yang, Yiming and Chen, Jiecao},
  journal={arXiv preprint arXiv:2510.11967},
  year={2025}
}

@misc{openai2026gpt54,
  author = {{OpenAI}},
  title  = {Introducing {GPT-5.4}},
  year   = {2026},
  month  = {March},
  howpublished = {\url{https://openai.com/index/introducing-gpt-5-4/}},
  note   = {Accessed: 2026-07-27}
}

@article{sellergren2026medgemma,
  title={Medgemma 1.5 technical report},
  author={Sellergren, Andrew and Gao, Chufan and Mahvar, Fereshteh and Kohlberger, Timo and Jamil, Fayaz and Traverse, Madeleine and Tono, Alberto and Sadjad, Bashir and Yang, Lin and Lau, Charles and others},
  journal={arXiv preprint arXiv:2604.05081},
  year={2026}
}

@article{simeoni2025dinov3,
  title={Dinov3},
  author={Sim{\'e}oni, Oriane and Vo, Huy V and Seitzer, Maximilian and Baldassarre, Federico and Oquab, Maxime and Jose, Cijo and Khalidov, Vasil and Szafraniec, Marc and Yi, Seungeun and Ramamonjisoa, Micha{\"e}l and others},
  journal={arXiv preprint arXiv:2508.10104},
  year={2025}
}

@article{zhang2025multimodal,
  title={A multimodal biomedical foundation model trained from fifteen million image--text pairs},
  author={Zhang, Sheng and Xu, Yanbo and Usuyama, Naoto and Xu, Hanwen and Bagga, Jaspreet and Tinn, Robert and Preston, Sam and Rao, Rajesh and Wei, Mu and Valluri, Naveen and others},
  journal={Nejm Ai},
  volume={2},
  number={1},
  pages={AIoa2400640},
  year={2025},
  publisher={Massachusetts Medical Society}
}

@article{oquab2024dinov2,
  title={Dinov2: Learning robust visual features without supervision},
  author={Oquab, Maxime and Darcet, Timoth{\'e}e and Moutakanni, Th{\'e}o and Vo, Huy and Szafraniec, Marc and Khalidov, Vasil and Fernandez, Pierre and Haziza, Daniel and Massa, Francisco and El-Nouby, Alaaeldin and others},
  journal={Transactions on Machine Learning Research Journal},
  year={2024}
}

@article{liu2024deepseek,
  title={Deepseek-v3 technical report},
  author={Liu, Aixin and Feng, Bei and Xue, Bing and Wang, Bingxuan and Wu, Bochao and Lu, Chengda and Zhao, Chenggang and Deng, Chengqi and Zhang, Chenyu and Ruan, Chong and others},
  journal={arXiv preprint arXiv:2412.19437},
  year={2024}
}

@article{zhang2019bertscore,
  title={Bertscore: Evaluating text generation with bert},
  author={Zhang, Tianyi and Kishore, Varsha and Wu, Felix and Weinberger, Kilian Q and Artzi, Yoav},
  journal={arXiv preprint arXiv:1904.09675},
  year={2019}
}

@inproceedings{lin2004rouge,
  title={Rouge: A package for automatic evaluation of summaries},
  author={Lin, Chin-Yew},
  booktitle={Text summarization branches out},
  pages={74--81},
  year={2004}
}

@article{teamqwen3,
  title={Qwen3. 5: Towards native multimodal agents (2026)},
  author={Team, Qwen},
  journal={URL https://qwen. ai/blog}
}

@article{xu2024whole,
  title={A whole-slide foundation model for digital pathology from real-world data},
  author={Xu, Hanwen and Usuyama, Naoto and Bagga, Jaspreet and Zhang, Sheng and Rao, Rajesh and Naumann, Tristan and Wong, Cliff and Gero, Zelalem and Gonz{\'a}lez, Javier and Gu, Yu and others},
  journal={Nature},
  volume={630},
  number={8015},
  pages={181--188},
  year={2024},
  publisher={Nature Publishing Group UK London}
}

@article{xu2025cardiac,
  title={A Cardiac-specific CT Foundation Model for Heart Transplantation},
  author={Xu, Hanwen and Woicik, Addie and Asadian, Sanaz and Shen, Junbo and Zhang, Zhengyan and Nabipoor, Ali and Musi, J Peter and Keenan, Jeffrey and Khorsandi, Maziar and Al-Alao, Bassel and others},
  journal={medRxiv},
  pages={2025--08},
  year={2025},
  publisher={Cold Spring Harbor Laboratory Press}
}

@article{liu2026three,
  title={A three-dimensional multi-modal foundation model for optical coherence tomography},
  author={Liu, Zixuan and Xu, Hanwen and Woicik, Addie and Shapiro, Linda G and Blazes, Marian and Wu, Yue and Steffen, Verena and Cukras, Catherine A and Lee, Cecilia S and Zhang, Miao and others},
  journal={Nature Biomedical Engineering},
  pages={1--20},
  year={2026},
  publisher={Nature Publishing Group UK London}
}

@article{usuyama2026gigapath,
  title={GigaPath-Flash and GigaTIME-Flash: Efficient Pathology Foundation Models for Whole-Slide and Tumor Microenvironment Analysis},
  author={Usuyama, Naoto and Valanarasu, Jeya Maria Jose and Yao, Sicong and Xu, Hanwen and Bagga, Jaspreet and Qin, Guanghui and Kramer, Robert E and Wong, Cliff and Lee, Soohee and Qiu, Hao and others},
  journal={arXiv preprint arXiv:2607.18218},
  year={2026}
}

@article{gao2025survey,
  title={A survey of self-evolving agents: On path to artificial super intelligence},
  author={Gao, Huan-ang and Geng, Jiayi and Hua, Wenyue and Hu, Mengkang and Juan, Xinzhe and Liu, Hongzhang and Liu, Shilong and Qiu, Jiahao and Qi, Xuan and Wu, Yiran and others},
  journal={arXiv e-prints},
  pages={arXiv--2507},
  year={2025}
}
\clearpage
\appendix
\section{Supplementary Notes}
\definecolor{promptbreak}{HTML}{9AA5B1}

\lstdefinestyle{promptstyle}{
  basicstyle=\ttfamily\footnotesize,
  breaklines=true,
  breakatwhitespace=true,
  breakautoindent=true,
  postbreak=\mbox{\textcolor{promptbreak}{$\hookrightarrow$}\space},
  columns=fullflexible,
  keepspaces=true,
  showstringspaces=false,
  frame=none,
  aboveskip=0pt,
  belowskip=0pt,
  xleftmargin=0pt,
  framexleftmargin=0pt,
}
\vspace{4pt}
\begin{lstlisting}[style=promptstyle]
Based on the echocardiogram pathology report content given, estimate the Aortic Stenosis - Indications for Surgery, look for the three indications as listed below:

1). Severe Aortic Stenosis: including but not limited to those with symptoms such as dyspnea, angina, syncope.
- Look for expressions like:
  - Severe aortic stenosis with [dyspnea/angina/syncope]
  - Patient with severe AS
  - Symptoms consistent with severe AS: [shortness of breath, chest pain, syncope]
  - Severe AS associated with exertional symptoms
  - Note : Symptom documentation may be sparse in the structured echo reports we provide. Symptoms are more likely found in clinical notes.

2). Severe AS with Low Flow, Low Gradient, Preserved EF: Severe AS with low flow (<4.0 m/s) and low gradient (<40 mmHg) when stroke volume index is <35 ml/m2 with preserved EF (EF>50%)
- Look for expressions or combinations of:
  - Aortic valve velocity < 4.0 m/s or peak velocity below threshold
  - Mean gradient < 40 mmHg
  - Stroke volume index < 35 ml/m2
  - Ejection fraction >= 50%, or normal left ventricular systolic function
  - Paradoxical low-flow, low-gradient AS
  - Severe AS with preserved EF and reduced stroke volume

3). Severe AS with Left Ventricular Systolic Dysfunction: Severe AS with low-flow and low gradient with left ventricular systolic dysfunction (EF < 50%)
- Look for expressions or combinations of:
  - Severe AS with low-flow, low-gradient in setting of reduced LV function (e.g., Mean gradient < 40 mmHg, velocity < 4.0 m/s, EF reduced)
  - EF measured at __% (below 50%), consistent with systolic dysfunction
  - Valve area critically reduced, but gradients are low due to impaired contractility
  - Dobutamine stress echo recommended to assess severity and contractile reserve
  - True severe AS likely despite deceptively low hemodynamic values
  - Severe AS with impaired LV function

Meanwhile, please also take care of the following CLEAR indications of NO surgery. When you see these conditions, mark the case as "no indication for surgery":

1). Mild or Moderate Aortic Stenosis (No Symptoms)
  - Explicitly described as mild or moderate AS
  - Peak velocity < 3.0 m/s or mean gradient < 20 mmHg
  - Valve area > 1.5 cm2
  - No dyspnea, angina, syncope, or other AS-related symptoms

2). Severe AS but Asymptomatic, Preserved LV Function, and No High-Risk Features
  - Severe AS without symptoms and normal exercise tolerance
  - EF >= 50%, normal stroke volume index
  - No abnormal blood pressure response on stress testing
  - No rapid progression or very high velocity (>5.0 m/s)

3) Severe AS with Major Contraindications to Surgery (non-valve factors)
  - Advanced comorbidities with limited life expectancy (<1 year)
  - Prohibitive surgical risks (e.g., severe frailty, active infection, severe non-cardiac illness)
  - Patient declines surgery after shared decision-making

Please follow the requirements below:

1. Create 4 attribute-value pairs
(1) surgery_indication_stated: <extracted value as stated in the pathology report>
(2) surgery_indication_estimated: <choose from -1, 0, 1, estimated based on the report and recent medical standards on echocardiogram, 0 for no indication for surgery, 1 for clear indication for surgery, and -1 for unknown (-1 means need further evidence from other resources such as observation or clinical notes etc.)>
(3) surgery_indication_confidence: <the certain degree of your estimation, within range [0.0, 1.0]>
(4) surgery_indication_evidence: <the supporting evidence of your estimation>

2. if there is no evidence please state "N/A" in surgery_indication_evidence, otherwise please list ALL the evidences that support your conclusion

3. please be sure to output all values together as a json string, and do not include anything beyond the json string, not even ```json and ```

4. if the Aortic Valve is not mentioned in the report, do not guess Aortic Stenosis, be precise and mark it as -1

5. please be careful to check whether or not the echo report is for a COMPLETE echo, if it is for a LIMITED echo, then not-stating anything does not necessarily means normal.

6. kindly provide careful and precise estimates, and use -1 for any cases where you're unsure

Now the report content is:
"{}"

What is your answer?
\end{lstlisting}

\captionof{figure}{\textbf{Prompt used to extract the aortic valve surgery indication label.} The prompt states the guideline criteria to determine surgery indications. It asks a large language model to read a clinical report and output a bare JSON object with four fields, including the verbatim supporting text, the extracted label, the confidence score and supporting evidence.}
\label{sfig:supp_surgery_prompt}
\newpage
\begin{lstlisting}[style=promptstyle]
Compare the following two echocardiogram reports and generate a concise longitudinal analysis. Focus on changes in cardiac structure, valve function, pressures, and systolic performance. Use clear clinical language and quantify differences where possible. Format the output as:

"Compared to the previous study (include date/time), the following changes were observed:"
1. Left Ventricular Size and Function: [Describe any notable changes in dilation, ejection fraction, or wall motion]
2. Valvular Findings: [Summarize changes in regurgitation, stenosis, gradients, or morphology]
3. Right Heart Parameters: [Note alterations in right ventricular size, function, or wall thickness]
4. Hemodynamic Metrics: [Include changes in pressures such as CVP, pulmonary artery pressure, or valve gradients]
5. Other Clinically Relevant Findings: [Mention any additional significant changes, such as pericardial effusion, chamber enlargement, or structural abnormalities]

Exclude any category where no meaningful change is identified or where findings are inconclusive.

Here's an example: "{}"

Please follow the requirements below:

1. Create 4 attribute-value pairs
(1) longitudinal_analysis_stated: <extracted longitudinal analysis as stated in the pathology report, if any; MUST be identical to the text from the report>
(2) longitudinal_analysis_estimated: <extimate a reasonable paragraph of longitudinal analysis, following the previous instructions>
(3) longitudinal_analysis_confidence: <the certain degree of your estimation, within range [0.0, 1.0]>
(4) longitudinal_analysis_evidence: <the supporting evidence of your estimation>

2. if there is no evidence please state "N/A" in longitudinal_analysis_evidence, otherwise please list ALL the evidences that support your conclusion

3. please be sure to output all values together as a json string, and do not include anything beyond the json string, not even ```json and ```

4. please be careful to check whether or not the echo report is for a COMPLETE echo, if it is for a LIMITED echo, then not-stating anything does not necessarily means normal

5. longitudinal analysis shouldn't be very long.

6. kindly provide careful and precise estimates, do not mention any information that you are unsure of.

Now the first report content is:
"{}"
and the second report content is:
"{}"

What is your answer?
\end{lstlisting}
\captionof{figure}{\textbf{Prompts used to construct the ground-truth
longitudinal report.} This prompt takes source reports of two consecutive echo studies as inputs, and describe the interval changes of abnormalities in the result.}
\label{sfig:supp_longitudinal_prompt}
\newpage

\vspace{4pt}

\begin{lstlisting}[style=promptstyle]
Compare the PREDICTED echocardiogram report to the GROUND TRUTH for the following {n_tasks} diagnostic tasks.

GROUND TRUTH:
{gt}

PREDICTED:
{pred}

For each task identify what each report says (or "not mentioned" if absent), then assign a deduction using these rules:
  0.0 - prediction is consistent with ground truth
  0.1 - prediction is close / slightly inaccurate
  0.2 - prediction differs noticeably from ground truth
  0.5 - prediction contradicts ground truth
  0.5 - ground truth does NOT mention this finding but prediction reports an ABNORMAL finding
  0.1 - ground truth does NOT mention this finding but prediction reports a NORMAL/mild finding

Tasks:
{tasks_json}

Respond with ONLY a valid JSON array, one object per task, in the same order:
[
  {"task": "<name>", "gt_finding": "<what GT says or 'not mentioned'>", "pred_finding": "<what pred says or 'not mentioned'>", "deduction": <0.0|0.1|0.2|0.5>, "reasoning": "<one sentence>"},
  ...
]
\end{lstlisting}
\vspace{6pt}

\captionof{figure}{\textbf{Prompt used for the LLM-as-a-judge evaluation of
single-study report factual accuracy.}}
\label{sfig:supp_judge_report_prompt}
\newpage
\vspace{4pt}

\begin{lstlisting}[style=promptstyle]
You are a strict cardiology evaluator grading an AI-generated LONGITUDINAL echocardiogram report against an expert ground-truth report. Both reports describe the INTERVAL CHANGE between a previous and a current study (improved / worsened / stable / new / resolved per cardiac dimension).

Use exactly this fixed dimension vocabulary:
  LV_function, LV_size, LV_hypertrophy, LA_size, RV_function, RV_size, RA_size, effusion, device_structure, other
(LV_function = LV systolic function / EF; effusion = pericardial/pleural effusion; device_structure = valve STRUCTURE, prosthesis/TAVR, calcification, vegetation, masses, thrombus, pacing leads/devices.)

STEP 1 - For every dimension the GROUND TRUTH mentions, determine its true direction (improved/worsened/stable/new/resolved) and magnitude (trace/mild/moderate/severe/marked/slight/null).
STEP 2 - For each such GT dimension, find the matching claim in the PREDICTION and assign ONE label:
  - 'correct'   : direction matches AND magnitude/severity matches (incl. correctly 'stable').
  - 'mag_off1'  : direction matches, magnitude off by ONE grade (e.g. mild vs moderate).
  - 'mag_off2'  : direction matches, magnitude off by >=2 grades, or pred gives no magnitude.
  - 'undercall' : one report asserts a change and the other says stable/unchanged (NO opposite direction is asserted) -- under- or over-call.
  - 'missed'    : the prediction does not address this dimension at all.
  - 'inverted'  : the prediction asserts the OPPOSITE direction (improved<->worsened, or new<->resolved). This is the worst error.
  Decision rule: if both reports assert a direction and they are opposite -> 'inverted'. If exactly one asserts a change and the other says stable -> 'undercall'. Never call it 'inverted' when one side merely says stable.

STEP 3 - List structural claims the PREDICTION makes that the GROUND TRUTH does NOT support (hallucinations); mark structural=true for new anatomic findings (thrombus, mass, vegetation, effusion, device, prosthesis) vs false for minor wording.

Return STRICT JSON only, no prose, no code fences:
{
  "per_dimension": [
    {"dimension": <one of the vocab>, "gt_direction": "...", "gt_magnitude": "...", "pred_direction": "...|absent", "label": "correct|mag_off1|mag_off2|undercall|missed|inverted", "reason": "<short>"}
  ],
  "hallucinated_claims": [{"text": "...", "structural": true|false}]
}
Only include dimensions the GROUND TRUTH actually mentions. Do NOT output a score; the harness computes scores from your labels.
\end{lstlisting}
\vspace{6pt}

\captionof{figure}{\textbf{Prompt used for the LLM-as-a-judge evaluation of
longitudinal report factual accuracy.}}
\label{sfig:supp_judge_longitudinal_prompt}

\newpage

\newpage

\newpage

\begin{table}[htbp]
\centering
\small
\begin{tabular}{llrrr}
\toprule
Abbr. & Abnormality & \# Train & \# Val & \# Test \\
\midrule
\multicolumn{5}{l}{\textit{Ventricular and atrial function}} \\
LHF & Left heart failure & 6{,}863 (6{,}679) & 375 (372) & 380 (374) \\
RHF & Right heart failure & 6{,}197 (5{,}903) & 333 (320) & 358 (344) \\
DF & Diastolic dysfunction & 4{,}087 (3{,}797) & 213 (202) & 224 (210) \\
LVD & Left ventricular dilation & 6{,}817 (6{,}518) & 380 (371) & 384 (365) \\
RVD & Right ventricular dilation & 6{,}213 (5{,}887) & 335 (322) & 358 (333) \\
LAD & Left atrial dilation & 5{,}933 (5{,}553) & 321 (304) & 339 (319) \\
RAD & Right atrial dilation & 5{,}797 (5{,}395) & 311 (287) & 333 (311) \\
LVH & Left ventricular hypertrophy & 5{,}630 (5{,}127) & 302 (276) & 308 (279) \\
TAPSE & Tricuspid annular plane systolic excursion & 4{,}453 & 225 & 258 \\
\addlinespace
\multicolumn{5}{l}{\textit{Valvular regurgitation and stenosis}} \\
AR & Aortic valve regurgitation & 5{,}910 (5{,}549) & 309 (285) & 331 (312) \\
AS & Aortic valve stenosis & 5{,}545 (4{,}864) & 285 (262) & 307 (265) \\
MR & Mitral valve regurgitation & 5{,}778 (4{,}775) & 304 (251) & 330 (267) \\
MS & Mitral valve stenosis & 4{,}757 (4{,}091) & 257 (226) & 279 (230) \\
TR & Tricuspid valve regurgitation & 5{,}748 (4{,}705) & 304 (241) & 340 (277) \\
TS & Tricuspid valve stenosis & 4{,}433 (3{,}787) & 227 (197) & 264 (224) \\
PR & Pulmonic valve regurgitation & 4{,}984 (4{,}068) & 259 (209) & 286 (232) \\
PS & Pulmonic valve stenosis & 4{,}126 (3{,}489) & 214 (188) & 237 (200) \\
\addlinespace
\multicolumn{5}{l}{\textit{Valvular vegetations}} \\
AV veg. & Aortic valve vegetations & 2{,}215 & 111 & 116 \\
MV veg. & Mitral valve vegetations & 2{,}190 & 110 & 122 \\
TV veg. & Tricuspid valve vegetations & 2{,}153 & 105 & 117 \\
PV veg. & Pulmonic valve vegetations & 2{,}024 & 100 & 110 \\
\addlinespace
\multicolumn{5}{l}{\textit{Other}} \\
PE & Pericardial effusion & 6{,}546 (979) & 352 (45) & 365 (68) \\
IMT & Intima-media thickening & 2{,}803 & 153 & 160 \\
IS & Interventricular septal abnormality & 5{,}263 & 286 & 303 \\
\bottomrule
\end{tabular}
\caption{\textbf{The 24 abnormalities used for detection and severity grading.} Table showing the number of studies in each split for the abnormality detection task. In parentheses, the count indicates the number of studies with a four-level severity grade for that abnormality if not zero. Tricuspid annular plane systolic excursion is documented only as a graded measure.}
\label{tab:supp_abnormalities}
\end{table}

\begin{table}[htbp]
\centering
\small
\begin{tabular}{lll}
\toprule
Abbr. & Abnormality & Views \\
\midrule
\multicolumn{3}{l}{\textit{Chamber size and systolic function}} \\
LHF & Left heart failure & 2C, 3C, 4C \\
RHF & Right heart failure & 4C, PSAX \\
LVD & Left ventricular dilation & 4C, 2C, 3C, PLAX \\
RVD & Right ventricular dilation & 4C, PSAX \\
LAD & Left atrial dilation & 4C, 2C, PLAX \\
RAD & Right atrial dilation & 4C \\
LVH & Left ventricular hypertrophy & PLAX, PSAX \\
TAPSE & Tricuspid annular plane systolic excursion & 4C \\
DF & Diastolic dysfunction & 4C \\
\addlinespace
\multicolumn{3}{l}{\textit{Valvular regurgitation and stenosis}} \\
AR & Aortic valve regurgitation & 3C, PLAX \\
AS & Aortic valve stenosis & PSAX, 5C \\
MR & Mitral valve regurgitation & 4C, 2C \\
MS & Mitral valve stenosis & 4C, PLAX \\
TR & Tricuspid valve regurgitation & 4C \\
TS & Tricuspid valve stenosis & 4C \\
PR & Pulmonic valve regurgitation & PSAX \\
PS & Pulmonic valve stenosis & PSAX \\
\addlinespace
\multicolumn{3}{l}{\textit{Valvular vegetations}} \\
AV veg. & Aortic valve vegetations & PLAX, 3C \\
MV veg. & Mitral valve vegetations & PLAX, 4C \\
TV veg. & Tricuspid valve vegetations & 4C \\
PV veg. & Pulmonic valve vegetations & PSAX \\
\addlinespace
\multicolumn{3}{l}{\textit{Other}} \\
PE & Pericardial effusion & PLAX, 4C \\
IMT & Intima-media thickening & 4C, 2C \\
IS & Interventricular septal abnormality & 4C \\
\bottomrule
\end{tabular}
\caption{\textbf{Map from abnormality to echocardiographic view.} Each of the 24 abnormalities is examined using the listed views. In \ourmethod, each tool call receives only the videos of corresponding views that an abnormality is mapped to. }
\label{tab:supp_task_views}
\end{table}

\end{document}